\documentclass{article}

    \usepackage[preprint]{neurips_2022}

\usepackage[T1]{fontenc}
\usepackage[utf8]{inputenc}
\usepackage{textcomp}

\usepackage[english]{babel}
\usepackage{xspace}
\usepackage{setspace}
\usepackage[final]{microtype}
\usepackage{balance}

\usepackage{float}
\usepackage[colorinlistoftodos]{todonotes}
\usepackage{wrapfig}
\usepackage{caption}
\usepackage{subcaption}
\usepackage{amssymb}
\usepackage{makecell}
\usepackage{amsmath, amsfonts, amssymb, nicefrac, pifont, siunitx}

\usepackage{algorithmic}

\DeclareMathAlphabet{\mathbbold}{U}{bbold}{m}{n}
\usepackage{bbm}

\usepackage{booktabs, array, tabu, multirow, multicol}
\usepackage[flushleft]{threeparttable}

\usepackage[inline]{enumitem}
\usepackage{listings}
\usepackage{colortbl}

\usepackage[hyphens]{url}
\usepackage[labelfont=bf]{caption}
\usepackage{hyperref}
\usepackage{natbib}

\usepackage[]{cleveref} %
\crefformat{section}{\S#2#1#3}
\crefrangeformat{section}{\S#3(#1)#4--\S#5(#2)#6}
\crefmultiformat{section}{\S#2#1#3}{ and~\S#2#1#3}{, \S#2#1#3}{, and~\S#2#1#3}
\crefformat{subsection}{\S#2#1#3}
\crefmultiformat{subsection}{\S#2#1#3}{, and~\S#2#1#3}{, \S#2#1#3}{, and~\S#2#1#3}
\crefformat{figure}{Fig.~#2#1#3}
\crefmultiformat{figure}{Figs.~#2#1#3}{, and~#2#1#3}{, #2#1#3}{, and~#2#1#3}
\crefformat{table}{Table~#2#1#3}
\crefrangeformat{table}{Tables~#3#1#4--#5#2#6}
\crefmultiformat{table}{Tables~#2#1#3}{, and~#2#1#3}{, #2#1#3}{, and~#2#1#3}
\crefformat{equation}{Eqn. #2#1#3}
\crefmultiformat{equation}{Eqns.~#2#1#3}{, and~#2#1#3}{, #2#1#3}{, and~#2#1#3}

\newcommand{\score}{{STI}\xspace}

\title{Watch Out for the Safety-Threatening Actors: Proactively Mitigating Safety Hazards}

\newcommand*\samethanks[1][\value{footnote}]{\footnotemark[#1]}

\author{%
  Saurabh Jha\thanks{Equal contribution.}  
   \And
  Shengkun Cui\samethanks[1]   
\space 
   \\
   \And
   Zbigniew T. Kalbarczyk
   \And
   Ravishankar K. Iyer
}

\begin{document}

\maketitle

\begin{abstract}
Despite the successful demonstration of autonomous vehicles (AVs), such as self-driving cars, ensuring AV safety remains a challenging task.
Although some actors influence an AV's driving decisions more than others, current approaches pay equal attention to each actor on the road.
An actor's influence on the AV's decision can be characterized in terms of its ability to decrease the number of safe navigational choices for the AV. 
In this work, we propose a safety threat indicator (\score) using counterfactual reasoning to estimate the importance of each actor on the road with respect to its influence on the AV's safety. 
We use this indicator to \begin{enumerate*}[label=(\roman*)]
    \item characterize the existing real-world datasets to identify rare hazardous scenarios as well as the poor performance of existing controllers in such scenarios; and 
    \item design an RL based safety mitigation controller to  proactively mitigate the safety %
    hazards those actors pose to the AV. Our approach reduces the accident rate for the state-of-the-art AV agent(s) in rare hazardous scenarios by more than 70\%. 
\end{enumerate*}
\end{abstract}

\section{Introduction}
\label{ml4ad:s:introduction}
Driving in a dynamic, real-world environment among other actors on the road is inherently a risky task, especially for autonomous vehicles (AVs). 
Each actor in the environment can significantly threaten the safety of an AV by (i) influencing the AV's driving decisions and (ii) limiting the number of the AV's safe trajectories. 
In extreme cases, any of these actors can, willingly (e.g., through adversarial attacks~\cite{chen2018shapeshifter, jha2020ml, jia2020fooling} or unwillingly (e.g., through accidents~\cite{banerjee2018hands, teslaCrash, uberCrash}), thwart the AV's ability to complete its driving task safely.
Humans maintain their safety by continuously monitoring the environment, identify the most safety threatening actors and reducing threats posed by taking preventive actions. 
However, current approaches to developing self-driving agents use either an end-to-end agent (e.g., learning by cheating (LBC) agent~\cite{chen2019lbc}) or a multipart agent composed of several deep learning models and heuristics (e.g., Apollo Baidu~\cite{apollo}). Both approaches focus on navigation and collision avoidance by minimizing perception/planning/control errors~\cite{chen2019lbc,pmlr-v119-filos20a,Philion_2020_CVPR}, rather than explicitly maximizing safety. 
Our evaluation of the state-of-the-art approaches shows that autonomous agents have a high accident rate (more than 50\%) in rare hazardous driving scenarios (refer to \cref{tab:miti_results}). 
Thus, there is need to (i) inculcate the human thinking of maintaining safety in AV by identifying and quantifying the threat posed by each actor in terms of their ability to create safety hazard, (ii) critique an AV agent's decisions to understand whether that agent is actively maximizing safety, and (iii) proactively mitigate those threats by ensuring that safe navigational choices are always available to the AV. 

\textbf{Our approach.}
We leverage the existing domain knowledge on how humans maintain safety.
Humans pay significant attention towards an actor if their presence (or actions) significantly reduces the number of safe navigational choices (thus decreasing backup safe choices), especially in hazardous scenarios in which there is limited reaction time.
Driven by this insight, we propose and formalize:
\begin{enumerate}[noitemsep,nolistsep,leftmargin=*, label=(\roman*), topsep=0pt]
\item safety threat indicator (\score), which is directly proportional to a reduction in the number of safe navigational choices available to the ego actor (the AV) in the presence of other nonplayer character (NPC) actors.
We assert (and empirically show) that reducing \score increases the number of safe navigational choices, thereby creating a safety blanket around the AV.
Finally, we estimate \score by causally modeling the problem and doing a counterfactual inference, which is ``given the observations, how does the absence of an NPC actor increases the number of safe navigational choices, and thereby, improves safety''. 

\item a reinforcement learning (RL) based safety hazard mitigation controller (SMC) that uses \score to proactively creates a safety blanket around an AV by reducing the \score using mitigation actions like braking (including emergency braking), acceleration, and lane change.
While there may be trade offs between passenger comforts or mission completion times and achieving this level of safety by reducing \score, 
our SMC learns to balance these trade-offs using multipart reward functions. 
The SMC execution is independent of the existing AV controller, so it does not interfere with the normal operation of the AV. 
This independent SMC approach 
alleviates the need to retrain (or redesign) the ego actor to support the additional mitigation actions.
\end{enumerate}

\begin{figure}[t]
    \centering
    \includegraphics[width=1.0\textwidth]{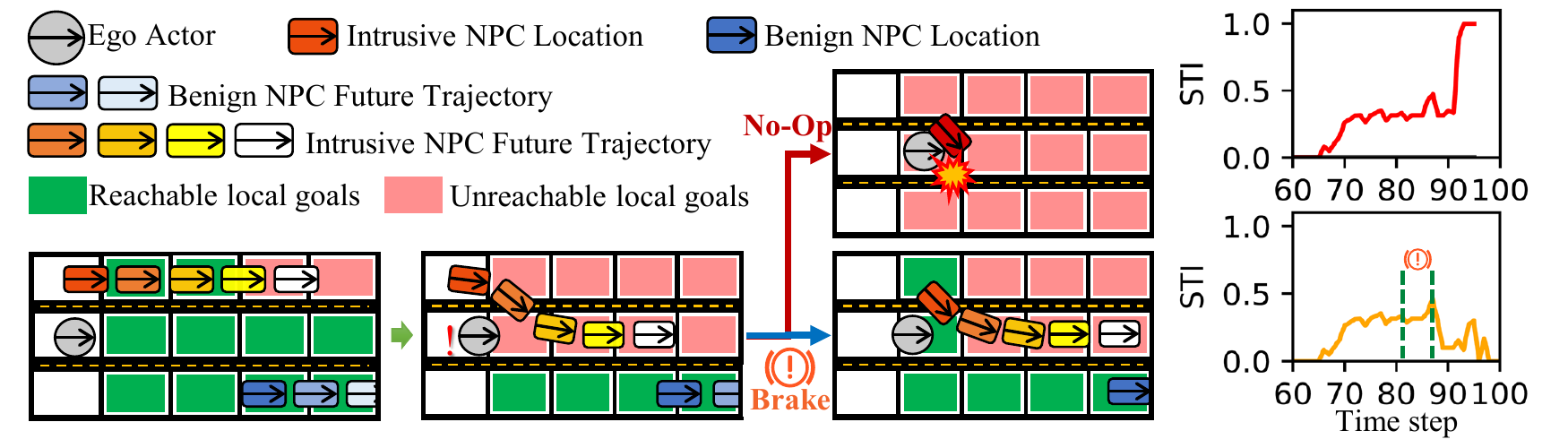}
    \caption{Safety threat indicator (\score) for a cut-in driving scenario with (a) safety mitigation controller (SMC) disabled (No-Op), and (b) SMC enabled (which decides to brake). 
    Each reachable grid (local goal location) represents a safe navigational path (from current location to goal location) while the path to unreachable grids are unsafe paths.
    The goal of the SMC is to maximize the number of available safe navigational choice so as to create the safety blanket around the AV.}
    \label{fig:approach}
\end{figure}

\textbf{Contribution.}
\cref{fig:approach} conceptually describes the core idea of our paper.
The contributions are:
\begin{enumerate}[noitemsep,nolistsep,leftmargin=*, label=(\roman*), topsep=0pt]
	\item \textit{Formalizing \score.} 
	To the best of our knowledge, this is the first attempt at using counterfactual reasoning to characterize the importance of each actor with respect to safety of the AV. The results are then used to proactively mitigate the safety hazards by taking precautionary actions.

	\item \textit{NN-based \score evaluation for meeting realtime latency requirements.} 
	Estimating \score to meet latency deadlines is computationally intractable because \score requires reachability analysis which is P-SPACE hard~\cite{lavalle1998rapidly}. 
	To solve this problem, we develop a neural network (NN) based approximation technique for estimating \score that is up to 414\texttimes~faster than traditional, planner-based \cite{5509799} \score estimation while achieving over $90\%$ accuracy on trained data. Given the low latency, \score can be used for monitoring and actively critiquing an AV's decisions.
	\item \textit{Demonstration of \score on real world driving datasets.}
	We evaluate the proposed indicator on Argoverse dataset~\cite{Chang_2019_CVPR} to extract safety-threatening actors and driving scenarios. 
	For example, we find that {\it only} 0.257\% of the actors and 4.75\% of the driving scenarios have \score of $\ge 0.9$.
	\item \textit{Demonstration of SMC in simulation.} 
	We demonstrate SMC (along with \score estimation) on Carla simulator and show that SMC enables an AV to reduce the number of accidents from 565 to 128 out of 1000 driving scenarios, a $77.3\%$ reduction in the accident rate.

\end{enumerate}

\subsection{Related work} 

\textbf{Actor importance.}
There is an increased focus on developing metrics to identify important actors. 
~\cite{Philion_2020_CVPR, piazzoni2020modeling} focus on evaluating %
the impact of the perception subsystem's accuracy and uncertainty on the downstream tasks of driving. 
In particular, Philion et al.~\cite{Philion_2020_CVPR} design a planning-centric metric based on ablation methods to understand how uncertainty in actor states can lead to significantly different planning outcomes.
Such a metric can be used to understand the importance of each actor, but it is not entirely satisfactory because (i) it is limited to improving the perception subsystem rather than safety of the overall system and (ii) it is agent dependent. %
In contrast, %
\score has different goals:  (i) adaptability, i.e.,  \score is independent of the agent's architecture, (ii) safety-oriented, i.e.,  \score explicitly encodes the impact of an actor or a scene on AV's safe execution.

\textbf{Proactive maintaining of safety.}
There are mainly two categories of work here: (i) identifying the safe distance from other actors, assuming everyone follows the  ``duty-of-care'' policies, and (ii) identifying out-of-training-distribution (OOD) scenarios so as to plan for the worst case in order to avoid collisions. 
\textit{Duty-of-care approaches}, such as Safety Force Field (SFF)~\cite{SafetyForceField} and Responsibility-Sensitive Safety (RSS)~\cite{shalev2017formal}, are geared towards estimating the safe distance from other actors assuming that everyone follows rules of the road. 
Additionally, the goal is to identify the culprit in case of a safety hazard. 
\textit{OOD} models, such as \cite{pmlr-v119-filos20a}, are geared toward identifying driving scenarios in which the ego actor's future trajectory (i.e., plan) has significant variance. 
Such variance can be characterized using an ensemble of models or diverse data.  
Under high variance, the ego actor chooses to use the most pessimistic plan in order to avoid collisions.

None of the above approaches proactively try to ensure that
multiple safe navigational choices are available to the AV.
In contrast, our goal is to proactively create a safety blanket around the AV, by ensuring that the AV has multiple navigational choices.
Our approach is not dependent on the underlying agent used by the AV. This allows the AV's agent to focus on complicated navigating by obeying the rules of the road tasks while the SMC focuses on safety.

\section{\score Formulation}
\label{ml4ad:s:problem-def}
Here we propose and formalize the safety threat indicator (\score) that quantifies the threat posed by an actor (and the driving scene) using causal and counterfactual reasoning.
This indicator allows us to rank the actors and driving scene in terms of the danger they pose to the AV, thus allowing us to develop attention-based safety and mitigation techniques.
We characterize the \score of an actor in terms of the decrease in driving flexibility caused by that actor. 
Driving flexibility is defined as the number of unique driving trajectories (or actions) available to the ego actor up to time horizon $k$ in the future at each time step. 
Thus, the greater the decrease in driving flexibility caused by an actor, the higher that actor's \score. 
\paragraph{Determining \score in the oracle setting\label{s:sim_gt}.}
We will first formalize the \score for an oracle setting in which the past, current, and future states of the actors are known with certainty. 
The oracle setting is suitable for offline assessment on prerecorded datasets where the ground truth labels are available. 
Let us assume that (i) there are $N$ actors in the world including the ego actor, (ii) we denote the state of an actor $i$ at time $t$ by $x^{(i)}_t \in \mathbb{R}^{3}$, and (iii) we denote the trajectory (i.e., trace of an actor's state over time) from time $t$ to $t+k$ as $X^{(i)}_{t:t+k}$.
Let us further denote the set of trajectories of all actors except the ego actor from time $t$ to $t+k$ by $\mathbb{X}_{t:t+k} = \{X^{(1)}_{t:t+k}, ..., X^{(N-1)}_{t:t+k}\}$. 
Let us further assume that (iv) we have access to ground truth information of $\mathbb{X}_{t:t+k}$ and an oracle local planner $f_{planner}$ that uses the trajectories of all the other actors from time $t$ to $t+k$, $\mathbb{X}_{t:t+k}$, and the position of the ego actor at time $t$, $x^{ego}_t$, to generate a set of future trajectories $\mathbb{Z}_{t:t+k}$ that the ego actor can follow safely from time $t$ to $t+k$ while obeying all the rules of the road (\cref{ml4ad:eq:planner}). 
Since the oracle planner $f_{planner}$ is a theoretical model, a practical implementation of \cref{ml4ad:eq:planner} is  discussed later in \cref{ml4ad:s:design}.
\begin{equation}
	Z_{t:t+k} = f_{planner}(\mathbb{M}, \mathbb{X}_{t:t+k}, x^{ego}_{t})
	\label{ml4ad:eq:planner}
\end{equation}
We can now construct a counterfactual query using \cref{ml4ad:eq:planner} in which we want to estimate the effect of removing $i^{th}$ actor (to understand the importance of the $i^{th}$ on safety) or all actors (to understand the importance of the driving scene) on $\mathbb{Z}_{t:t+k}$.
The set consisting of all the navigable future trajectories in the absence of all actors, denoted $Z^{\varnothing}_{t:t+k}$, can be obtained using  \cref{ml4ad:eq:planner} with $\mathbb{X}_{t:t+k} = \varnothing$. 
Similarly, the set consisting of all navigable future trajectories in the absence of the $i^{th}$ actor, denoted $\mathbb{X}^{/i}_{t:t+k}$, can be obtained using \cref{ml4ad:eq:planner} with $\mathbb{X}_{t:t+k} = \mathbb{X}^{/i}_{t:t+k}$, where $\mathbb{X}^{/i}_{t:t+k}$ is the actor trajectory set that contains trajectory of all actors except the $i^{th}$ actor.

\noindent
We can now define the \score of a specific actor $i$ from time $t$ to $t+k$ as:
\begin{equation}
	\rho^{(i)}_{t:t+k} = \frac{|Z^{/i}_{t:t+k}| - |Z_{t:t+k}|}{|Z^{\varnothing}_{t:t+k}|} 
	\label{ml4ad:eq:actorrisk}
\end{equation}
where $|\cdot|$ denotes the cardinality of the set. 
$|Z^{\varnothing}_{t:t+k}|$ is the normalization constant that forces the range to $[0, 1]$ and enables us to compare \score across different driving scenarios. 
The \score value of $0$ means actor $i$ does not constraint the Ego actor's driving flexibility. \score value of $1$ means %
that actor $i$ %
constraints the Ego actor's driving flexibility such that there is no viable driving trajectory.

We can similarly define the total \score, $\rho_{t:t+k}$, of a driving scene per time step $t$ as the normalized reduction in future trajectories due to presence of all actors on the road. 
\begin{equation}
	\rho_{t:t+k} = \frac{|Z^{\varnothing}_{t:t+k}| - |Z_{t:t+k}|}{|Z^{\varnothing}_{t:t+k}|} 
	\label{ml4ad:eq:totalrisk}
\end{equation}
\paragraph{Determining \score in the realtime setting\label{s:sim_rt}.}
We now extend the formalism for realtime setting in which the precise locations of the actors are not known and in which one can only approximate the past, the current, and the future states of the actors using sensor measurements and predictive models, which include uncertainties.
In a realtime setting, instead of point estimates (as discussed in \cref{s:sim_gt}), we deal with probability distributions that allow us to model the noise. 
The complete derivation is provided in Supplementary Material.

\subsection{Implementation of \score for realtime setting}
\label{ml4ad:s:design}

Designing $f_{planner}$ is computationally intractable because the oracle planner is expected to output the set of all future safe trajectories (countable infinite), which is P-SPACE hard.
Therefore, we propose following two heuristics: (i) a relaxed formulation that allows us to compute \score using planners employed in practice such as RRT~\cite{lavalle1998rapidly}, FOT~\cite{5509799}, and Hybrid A*~\cite{dolgov2008practical} and (ii) a data-driven model that enables realtime \score evaluation.

\paragraph{\score evaluation using practical planners\label{s:reach_cal_fp}.}
Instead of calculating all viable future trajectories to the goal location, we propose to calculate reachable alternative locations (hereafter, referred to as the \textit{local goals}) near the ego vehicle as a proxy representation of driving flexibility.
These (local) goals are safe and represent intermediate locations that the ego vehicle can navigate to while trying accomplish its overall task of reaching the destination goal (global goal) safely.
The local goals are the individual cells within the threshold distance $d$ of the current ego actor's location in the bird's-eye-view (BEV) of the drivable area\footnote{We discretize the drivable area into a grid of cells; see Supplementary Material for visualization} that are potentially reachable given the time budget. 
Since $d$ has an upper bound given that the vehicle states,  dynamic specifications, and number of grids in the drivable areas are fixed, the set of goals is also finite. 
{This representation is a reasonable approximation of driving flexibility because the number of reachable local goals is positively correlated with the number of safe trajectories, and vice-versa}. 
We denote the set of local goals that are potentially reachable with in the next $k$ time steps ($k$ is the time budget) as ${G}_k$. We can rewrite \cref{ml4ad:eq:planner} as a reachability evaluation of the set of local goals ${G}_k$ given a reachability solver $f_{reach}$. The reachability solver $f_{reach}$ classifies each goal $g_k \in {G}_k$ as either reachable or not-reachable and returns the set of reachable local goals ${R}_k = \{g_k | g_k \in {G}_k, g_k \text{ is reachable between } t \text{ and } t+k\}$ given the time budget $k$ (\cref{eq:reach_planner}).
\begin{equation}
	{R}_k = f_{reach}(\mathbb{M}, \mathbb{X}_{t:t+k}, x^{ego}_{t}, {G}_k; \hat{f}_{planner}), {R}_k \subseteq {G}_k
	\label{eq:reach_planner}
\end{equation}
We can then rewrite \cref{ml4ad:eq:actorrisk} by replacing the set of safe trajectories $Z_{t:t+k}$ with the set of reachable local goals ${R}_k$ for \score evaluation as
\begin{equation}
	\rho^{(i)}_{t:t+k} = \frac{|R^{/i}_{k}| - |R_{k}|}{|R^{\varnothing}_{k}|} 
	\label{eq:reach_actorrisk}
\end{equation}
Similarly, we can rewrite \cref{ml4ad:eq:totalrisk} as
\begin{equation}
	\rho_{t:t+k} = \frac{|R^{\varnothing}_{k}| - |R_{k}|}{|R^{\varnothing}_{k}|} 
	\label{eq:reach_totalrisk}
\end{equation} with $R^{\varnothing}$ and $R^{/i}_{k}$ being the set of reachable local goals in the absence of all other actors and a particular actor $i$, following the same reasoning of $Z^{\varnothing}_{t:t+k}$ and $Z^{/i}_{t:t+k}$, respectively.
The reachability problem in \cref{eq:reach_planner} can be solved efficiently using a practical planner  $\hat{f}_{planner}$~\cite{lavalle1998rapidly, 5509799, dolgov2008practical} (treated as parameters of $f_{reach}$) given the current ego actor state $x^{ego}_{t}$, the states of the actors $\mathbb{X}_{t:t+k}$, the drivable areas $\mathbb{M}$, and the ego actor's dynamics capabilities (as part of the planner parameters). 
Depending on the problem setting, $\mathbb{X}_{t:t+k}$ can either be the ground truth actor states from a labeled dataset or the predicted states %
from the actors' state histories (recall \cref{s:sim_rt}).
\paragraph{\score approximation using a data-driven model\label{s:nn_reach}.}
The evaluation of \score using \cref{s:reach_cal_fp} requires reachability analysis per actor per goal.  Using a planner-based $f_{reach}$ to calculate the reachability of multiple local goals for multiple passes (each pass masks an actor) is not favorable because of (i) the computational complexity of the planner algorithm (P-SPACE hard despite the previous optimization) and (ii) the linear time complexity w.r.t. the number of local goals to be evaluate times the number of other actors in the scene. Since our ultimate goal is reachability evaluation, but not trajectory generation, to reduce \score evaluation latency we construct a neural network (NN) based reachability approximator (hereafter referred as the \textit{NN approximator}), denoted $\mathcal{N}_{reach}$, to approximate the reachability calculation of the $f_{planner}$ (the \textit{FOT} planner in our implementation). 
We choose to approximate reachability instead of the \score directly to preserve the interpretability of our formulation, i.e., help users of the model understand which local goals are reachable vs unreachable.

The input to $\mathcal{N}_{reach}$ consists of (i) a set of ego-actor-centered, bird's-eye-view (BEV) images $\mathbf{I}_{t:t+k}$, that capture all the essential spatial-temporal information from time $t$ to $t+k$ of the scene and (ii) the ego actor's state vector $\mathbf{e}_t$. 
$\mathbf{I}_{t:t+k} = \{I_t, I_{t+1}, ..., I_{t+k}\}$ is constructed by stacking the BEV images from time step $t$ to time step $t+k$. The BEV image of a time step $t$, denoted ${I}_{t}$, captures drivable areas and the actors' locations as spatial information. Stacking BEV images from multiple time steps allows the input data to capture the evolution of the actor's future trajectories over time as temporal information.
In addition, the current state of the ego actor, $\mathbf{e}_t = (v_{long}, a_{long}, v_{lat}, a_{lat})$, encodes the velocity and acceleration information for the current time step as velocity impacts how far the vehicle can travel in the next $k$ time steps. Since the reachability analysis uses an ego-centered coordinate frame, the location of the ego actor is not needed. 
Given $\mathbf{I}_{t:t+k}$ and $\mathbf{e}_t$ the model $\mathcal{N}_{reach}$ outputs the reachability indicator vector $\mathbf{r}_k$ for the current time step:
\begin{equation}
	\mathbf{r}_k = \mathcal{N}_{reach}(\mathbf{I}_{t:t+k}, \mathbf{e}_t)
	\label{ml4ad:eq:nn_planner}
\end{equation}
Each element of the reachability indicator vector $\mathbf{r}_k$ is a binary value indicating whether a local goal location in the input BEV image $I_t$ is reachable. Hence we formulate the reachability problem as a binary classification problem per $\mathbf{r}_k$ element, where the goal of the NN is to classify each cell in the grid (or each element of $\mathbf{r}_k$) as reachable verses not-reachable. Therefore, given a ground truth indicator vector label $\mathbf{r}^*_k$, the objective function to minimize is
\begin{equation}
	l(\theta) = \sum_{b \in \mathcal{B}} BCE(\mathbf{r}^{(b)}_k, \mathbf{r}^{*(b)}_k) = \sum_{b \in \mathcal{B}} BCE(\mathcal{N}_{reach}(\mathbf{I}^{(b)}_{t:t+k}, \mathbf{e}^{(b)}_t; \theta), \mathbf{r}^{*(b)}_k)
	\label{ml4ad:eq:nn_obj_func}
\end{equation}
where $BCE(\cdot)$ stands for binary cross-entropy loss and $\mathcal{B}$ is the mini-batch. The ground truth labels $\mathbf{r}^*_k$ can be generated using the $f_{reach}$ with $\hat{f}_{planner}$ following \cref{s:reach_cal_fp}.
Note that to ensure a fixed size $\mathbf{r}_k$, instead of discretizing the drivable area as in \cref{s:reach_cal_fp}, we discretize the entire BEV image space. We do this because the BEV image size is fixed, while the drivable area size can vary.
The detailed generation of $\mathbf{I}_{t:t+k}$ as well as $\mathbf{r}_k$ is provided in the Supplementary Material.
Finally, given the predicted reachability indicator vector $\mathbf{r}_k$, we can assess whether a goal location in the BEV image space is reachable and hence calculate the actors and the scene \score using \cref{eq:reach_actorrisk,eq:reach_totalrisk}.

\section{RL-based Mitigation for Safety-Critical Scenarios}
\label{s:rl_agent1}
Safety-critical scenarios are scenarios that can result in serious safety violations~\cite{teslaCrash, uberCrash,banerjee2018hands} if not
handled properly. 
Here we describe the model and architecture of our 
proactive safety-hazard mitigation controller
(SMC). 
SMC uses RL and \score to perform mitigation actions in safety-critical scenarios.
The SMC is a standalone module that executes alongside the ego actor and intervenes with mitigation actions
when necessary. 
The overall architecture of SMC is shown in \cref{fig:framework}.

To illustrate the execution model with the SMC, suppose we have a ego actor denoted as $\mathcal{A}$ that is capable of handling normal driving scenarios. At each time step $t$, $\mathcal{A}$ consumes the state $\mathbf{S}_t$ and perform a driving action $da_t$: $da_t = \mathcal{A}(\mathbf{S}_t)$. The SMC, denoted $\mathcal{M}$, also consumes some state $\mathbf{S'}_t$ (which can be the same as $\mathbf{S}_t$) and provides a mitigation action $ma_t$ based on its policy: $ma_t = \mathcal{M}(\mathbf{S'}_t)$. $ma_t$ overwrites $da_t$ if $ma_t$ is not the "no-operation" (No-Op) action.  So for the final action $a_t$ on $\mathbf{S}_t$: 
\begin{equation}
	a_t = da_t \text{ if } ma_t == \text{No-Op} \text{ else } a_t = ma_t
	\label{equ:action_mod}
\end{equation}

Next we define the state, action space, state transitions, and reward function used to learn the policy.

\textit{State space}: The state $\mathbf{S}_t \in \mathbb{S}$ at time $t$ captures the spatial-temporal information of the SMC's environment, including other actors and drivable areas. It could be the internal world of the ego actor or a direct sensor measurement such as camera images (\cref{s:exp_smc}) or LiDAR cloud points. The design can be flexible depending on what information is available.

\textit{Action space}: The action space $\mathbb{A}$ of the SMC consists of mitigation actions; this includes but is not limited to emergency-braking ($EB$), lane-change-to-left ($LCL$), lane-change-to-right ($LCR$), accelerate ($ACC$), etc., plus no-operation (No-Op). 
Note that No-Op is needed because it provides the SMC with the option of not performing mitigation in non-safety-critical situations. 
In this work the actions are discrete, in which case the actions' trajectories and low-level controls are predefined and the SMC selects which action to perform. 

\textit{State transition function}: The state transition function $f_{transit}: \mathbb{S} \times \mathbb{A} \rightarrow \mathbb{S}$ calculates the next state $\mathbf{S}_{t+1}$ given the current state $\mathbf{S}_{t}$ and action $ma_t$ pair: $\mathbf{S}_{t+1}$ = $f_{transit}(\mathbf{S}_t, ma_t)$. Here, $f_{transit}$ is provided by the driving simulator~\cite{carla17,lgsvl}. At each time step, the simulator provides $\mathbf{S}_t$ for the actor to act. The actor acts on $\mathbf{S}_t$ based on its policy, and the simulator upon receiving the action updates the scenario and provides $\mathbf{S}_{t+1}$.

\textit{Rewards}: At each time $t$, a reward $r_t$ is generated based on the state-action pair. To effectively learn a mitigation policy that minimizes the scenario \score over time, we use \score as part of the reward function. In addition, the reward function is constructed to encourage normal driving and to penalize mitigation actions in low \score situations. A generic reward is as follows:
	\begin{equation}
		r_t = \alpha_0 (1-STI) + \alpha_1 r_{pc} + \alpha_2 p_{am} + \alpha_3 r_{comfort} + ... + \alpha_n r_{nth\_factor}
		\label{ml4ad:eq:rl_eward}
	\end{equation}
	 where $\alpha_0 ... \alpha_n, \alpha_i \ge 0, \sum_{i=0}^n \alpha_i = 1$ are constant weights of each term %
	 that controls trade-offs. $STI, r_{pc}, p_{am}$, and $r_{comfort}$ are \score values of the scene, reward for path-completion, penalty for activation of mitigation, and reward for maintaining comfort, respectively. Note that the third term penalizes frequent activation of mitigation actions because safety-critical situations are rare and mitigation should only be activated when it is truly needed, to avoid affecting normal driving.  
	 The general formulation also shows additional performance indicator terms that can be customized for different use-cases. 
	 The definitions of $ r_{pc}$, $p_{am}$, and $r_{comfort}$ used in our implementation are found in Supplementary Material.

\section{Experiment Setup}
\begin{figure}[t]
	\centering
	\includegraphics[width=1.0\textwidth]{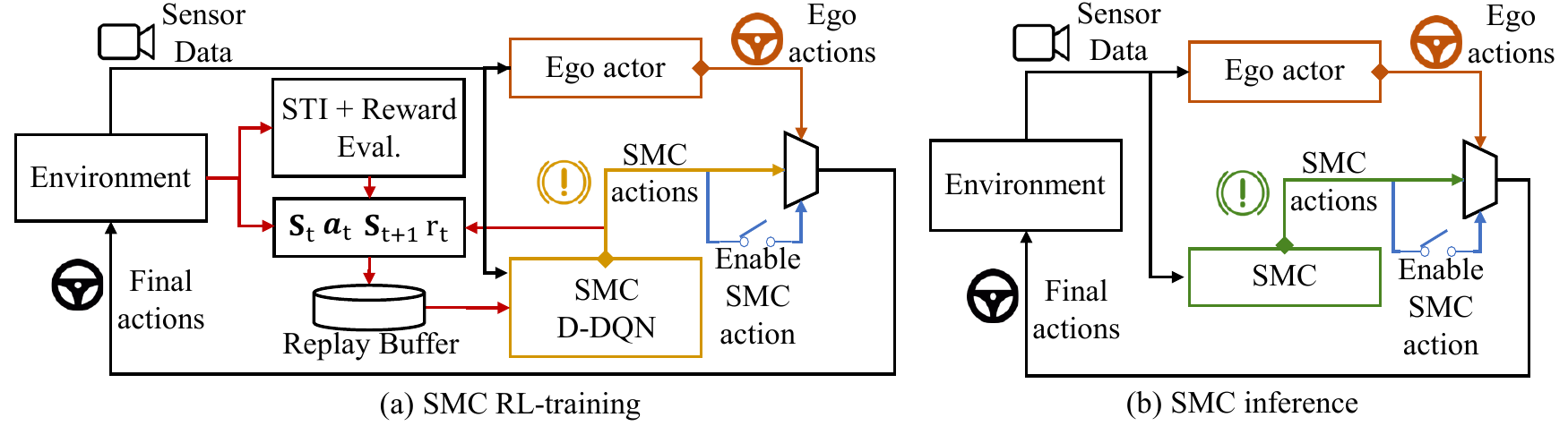}
	\caption{Overview of the SMC framework: (a) RL-based SMC training with \textit{Double DQN} and experience replay, (b) Runtime SMC inference}
	\label{fig:framework}
\end{figure}
This section describes in detail the dataset, training, and evaluation (also shown in \cref{fig:framework} ).
\subsection{Dataset and driving scenarios}
\label{s:datasets}
\paragraph{Real-world dataset.} 
We characterize the Argoverse real-world dataset~\cite{Chang_2019_CVPR} to evaluate the bias that exists in such datasets towards safer scenarios.
Such bias exists because the vehicles used in these datasets are controlled (or monitored) by humans who tend to avoid dangerous scenarios.  
The Argoverse dataset consists of $50+$ driving scenes, $300$K actor annotations, and HD-map information.
The Argoverse dataset is also used to train and evaluate the NN-based reachability approximator.
\paragraph{Simulator-generated dataset.}
The real-world datasets consist of prerecorded data that (i) do not provide a complete control-feedback loop and (ii) lack safety-critical scenarios with high actor or scene \score values (refer to \cref{ss:char-sti}).
Therefore, we use Carla~\cite{carla17} to simulate and collect data on safety-critical scenarios.
We simulate the cut-in scenario from the NHTSA precrash scenario typology report~\cite{najm2007pre} as the base scenario because it ranks among the top scenarios for financial loss and fatality rate. 
In the cut-in scenario, one of the NPC actors is approaching from behind the ego actor but in the adjacent lane, and then the NPC actor cuts in front of ego actor closely as it passes the ego actor.
Such a cut-in scenario is highly hazardous due to the limited reaction time available before collision. 
Note that there are multiple other NPC actors but those are not as hazardous as the ones that are cutting in.
The base scenario defines the overall scenario typology, but the level of safety criticality may vary depending on the scenario's hyper-parameters such as (i) cut-in distance, (ii) cut-in speed, (iii) cut-in angle and (iv) the distance between the ego actor and the NPC actor at the start of the scenario.
Therefore, to stress test our proposed methodology, we generate 1000 mutated scenarios based on the base scenario by varying the above hyperparameters (described in Supplementary Material). 
These mutated scenarios are used to (i) evaluate the correlation between the \score (refer to \cref{ml4ad:s:problem-def}) and the safety criticality of the scenarios, (ii) train and evaluate the NN-based reachability approximator, and (iii) train and evaluate the RL-based SMC. 
For each mutated scenario, we run the same ego actor (the LBC agent~\cite{chen2019lbc}). 
Hereafter, we will refer to the resulting dataset as the \textit{mutated scenarios dataset}.

\subsection{Training procedure}
\label{s:train_val}

\subsubsection{NN-based reachability approximator}
\textbf{Inputs and outputs:} We train and evaluate the approximator on both the real-world dataset and the simulated mutated scenarios dataset. 
We use the bird's-eye-view (BEV) to represent the drivable area along with the NPC actors as the input $\mathbf{I}_{t:t+k}$ to the approximator.
Note that we need the ground truth labels, in this case the reachability indication vector $\mathbf{r}_k$, to train the approximator using the loss function \cref{ml4ad:eq:nn_obj_func}. 
We generate these ground truth labels using \textit{FOT} planner~\cite{5509799} to calculate the reachability for each input $\mathbf{I}_{t:t+k}$. 
We choose the \textit{FOT} planner because it considers physically realistic trajectories given the ego actor's dynamics and the time horizon $k$ for the reachability calculation. %

\textbf{Model architecture:} The approximator uses the \textit{ResNet-34}~\cite{resnet34} as the feature extractor to extract information from the input $\mathbf{I}_{t:t+k}$ (see \cref{ml4ad:eq:nn_planner}). 
We replace the fully connect layers for classification with a customized output head to convert the extracted feature into the reachability indication vector $\mathbf{r}_k$ (see \cref{ml4ad:eq:nn_planner}). 
Please see the architectural details in the Supplementary Material.

\textbf{Training procedure:}
We do $80\%-20\%$ training-validation split to assess the approximator's accuracy.
We train the model for 750 epochs with per $\mathbf{r}_k$ element \textit{Binary Cross-entropy} loss as the criterion and the \textit{Adam Optimizer}~\cite{kingma2014adam} with a learning rate of $1e^{-5}$ to optimize the model parameters.

\subsubsection{RL-based SMC}
\label{s:exp_smc}
\textbf{Inputs and Outputs:}
The RL-based safety harzards mitigation controller (SMC) is implemented based on the design in \cref{s:rl_agent1}. 
The state representation $\mathbf{S}_t$ of the environment is the same as the input (three front-facing camera images) to the ego actor (\textit{LBC} agent) except that we convert the images to grayscale as the input instead of using the RGB format. 
To embed temporal information as part of the state representations, we follow the method outlined in~\cite{mnih2013playing}, in which the authors concatenate the grayscale camera data plus the planner target heatmaps from 4 previous time steps. 
At each time step $t$, the SMC $\mathcal{M}$ consumes the state and outputs a mitigation action $ma_t$ based on its mitigation policy. 

\textbf{Model architecture:}
The SMC follows ~\cite{mnih2013playing}, in which a CNN ($V$) with parameter $\phi$, denoted $V_{\phi}$, is used as the function estimator to approximate the $Q(S,a)$ values for all actions given $\mathbf{S_t}$: $$Q(a_t,\mathbf{S_t}; \phi) = V_{\phi}(\mathbf{S_t})$$
In our implementation, $V_{\phi}$ uses a similar backbone as the ego actor with less input channel due to the usage of grayscale input images. 
We modify the output head to output the same size as the action space $\mathbb{A}$ to predict the $Q$ values for all possible actions in one shot. At inference time, the action with the maximum $Q$ value is chosen.  
More details are available in Supplementary Material. 

\textbf{Training procedure:}
\label{s:smc_training}
We train the SMC on the simulated cut-in scenario by executing 100 episodes of a mutation with hyperparameters equal to the median value of the individual hyperparameters across the previously generated 1000 mutations of the cut-in scenario (which is described in ~\cref{s:datasets}). At each training time step, the mitigation actor (SMC) observes the state and select mitigation activation based on the learned policy. If mitigation is activated ($ma_t \ne \text{No-Op}$), the control output by the ego actor is overridden by the SMC's mitigation action (\cref{equ:action_mod}). 
Note that we do not modify the policy of the ego actor but only train the SMC. 
Since the reward function depends on the \score values evaluation per frame, in order to run the training closer to realtime, we use the CNN-approximator to evaluate the scene \score values per time step given its high accuracy (see \cref{s:nn_accuracy}). Finally, the Double-DQN~\cite{DBLP:journals/corr/HasseltGS15} training algorithm with replay buffer is used for faster convergence.

\section{Results}

\subsection{Characterization of \score values}
\label{ss:char-sti}
We characterize the \score of the actors and the scenes for both the real-world and the variant scenarios datasets in \cref{s:datasets}. 
The \textit{FOT} planner-based \score evaluator is used to characterize the \score value of all datasets in an offline setting to maximize accuracy. 

\subsubsection{Real-world dataset \label{s:real_wor_res}}
\begin{figure}[t!]
	\centering
	\subcaptionbox{~\label{subfig:real_actor_sti}}{\includegraphics[width=0.23\textwidth]{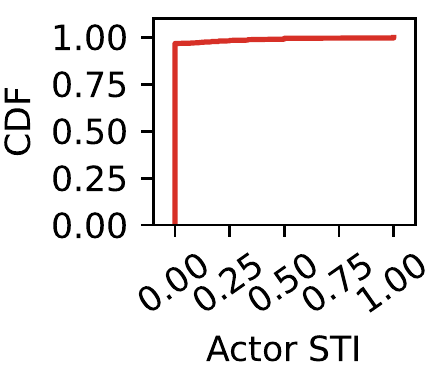}}
	\subcaptionbox{~\label{subfig:real_scene_sti}}{\includegraphics[width=0.23\textwidth]{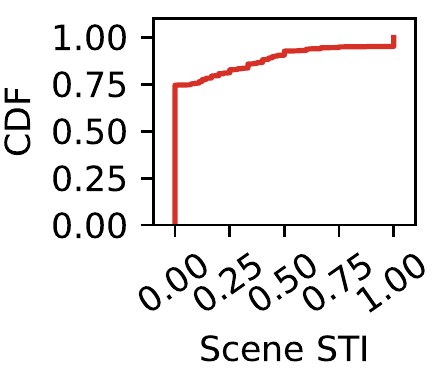}}
	\subcaptionbox{~\label{subfig:ghostcutin_boxplot}}{\includegraphics[width=0.23\textwidth]{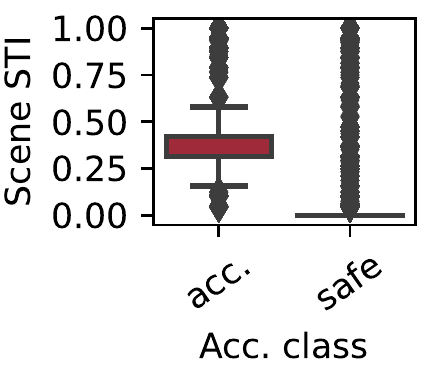}}
	\caption{\label{fig:realdata_sti} (a)(b) \score CDF of \textit{Argoverse} dataset: (a) Actor STI (b) Scene STI; (c) \score distribution vs accident class of the cut-in scenario; acc.: mutations with accidents, safe: mutations w/o accidents.}
\end{figure}
We evaluate the \score values of (i) each actor in a driving scene (per time step) and (ii) each driving scene (per time step) across all driving scenes in the \textit{Argoverse} dataset. 
According to the evaluation results, the $50^{th}$, $75^{th}$, $90^{th}$, and $99^{th}$ percentiles of the actor \score values are 0.0, 0.0, 0.0, and 0.4, respectively, while the $50^{th}$, $75^{th}$, $90^{th}$, and $99^{th}$ percentiles of the scene \score values are 0.0, 0.071, 0.454, and 0.99 respectively, as depicted in the CDFs of the \score value distribution in \cref{subfig:real_actor_sti,subfig:real_scene_sti}.This means that at least $90\%$ of the actors or $50\%$ of the scenes have 0 \score values and that a high \score is rare and belongs to the long tail of the real-world datasets' data distribution. This characterization exposes the current limitations of the state-of-the-art, real-world datasets when it comes to including a sufficient number of high \score actors and scenes for unbiased AV development as the datasets are biased towards safer scenarios; thus, it highlights the need for inclusion of such data and the need for a simulated dataset in cases where high \score actors and scenes can be generated in large numbers. Note that a scene's overall \score can be high even when the importance of the individual actors is low, as multiple actors together can significantly reduce the driving flexibility of the ego actor. We will explain such examples in the Supplementary Material. 

\subsubsection{Simulated safety-critical dataset}
\label{s:sc_dataset_gen}

We evaluate the \score values on the variant scenario dataset to understand the relationship between \score values and safety violations (accidents). 
\cref{subfig:ghostcutin_boxplot} shows the boxplot distribution of \score across all driving scenes per time step (calculated with \cref{eq:reach_totalrisk}) for runs with and without accident. 
The figure shows that \score values are significantly higher in the scenes that lead to accidents in the future compared to those without, suggesting that high \score and the chance of accident are positively correlated. 
Moreover, our analysis shows that the underlying distribution of \score is different between scenes that lead to accidents verses scenes that do not. This is verified by the statistical comparison using \textit{KS Two-sample Test} on the distributions of \score from scenes that lead to accidents verses the \score from scenes that are safe, with a significance level of $0.05$. We can reject the Null-hypothesis of the \textit{KS Two-sample Test}, as the resulted $p\text{-}value$ is \textless{}\textless $0.01$ (outputs 0.00 due to numerical underflow). 

\cref{subfig:ghostcutin_prob} shows the probability that an accident will happen within the next $k$ time steps\footnote{Recall $k$ from \cref{s:sim_gt}. Here we choose $k=30$, which equals 3 seconds in simulation time.} given a particular \score value for the cut-in scenario mutations.
We can see a positive relationship between the \score values and the accident probability in the next $k$ time steps, which suggests that \score values are good indicators of imminent safety violations. 

\subsection{Evaluation of the NN-based reachability approximator}
\label{s:nn_accuracy}
This section shows the evaluation results of the NN-based reachability approximator on both the real-world dataset and the variant scenarios dataset.

\begin{table}[]
\begin{subtable}[t]{0.48\textwidth}\centering
	\begin{tabular}[t]{l|l|l|l}
		\toprule
		 \textit{Dataset} & \makecell{\textbf{Precision}}  &  \makecell{\textbf{Recall}} &  \makecell{\textbf{F1-score}}      \\
		\midrule                             
		\textit{Argoverse}             & $0.83$                &  $0.87$             & $0.85$  \\ \hline
		\makecell{\textit{Mutated} \\ \textit{scenarios}}        & $0.96$                &  $0.97$             & $0.96$  \\
		\bottomrule
	\end{tabular}%
	\caption{\label{tab:approx_eval} NN-based reachability approximator performance on testing datasets (described in \cref{s:datasets}).}
\end{subtable}%
\hspace{\fill}
\begin{subtable}[t]{0.48\textwidth}\centering
	\begin{tabular}[t]{c|l}
		\toprule
		\textit{Mitigation policy}  & \makecell{\# acc./\# T.runs} \\  %
		\midrule                             
		\textit{\small{LBC}}                              &  565/1000 (0.565)   \\ \hline %
		\makecell{\small{\textit{LBC+SMC}} (ours)} &  \textbf{128/1000 (0.128)}   \\ %
		\bottomrule
	\end{tabular}%
	\caption{\label{tab:miti_results} Comparison of mitigation policies (i) \textit{LBC} agent and (ii)  \textit{LBC} agent with SMC (ours); \# acc.: \#runs with accident, \#T.runs: total runs.}
\end{subtable}%
\caption{\vspace{-0.2cm}Table of results: (a) NN-based reachability approximator evaluation, (b) SMC evaluation.}
\vspace{-0.25cm}
\end{table}

\cref{tab:approx_eval} summarizes the reachability approximation performance on both the real-world and the simulated datasets. 
The approximator performs worse for the real-world dataset because the HD-map information is not accurate compared to the simulator maps.

The NN-based approximator reduces the reachability compute latency on the real-world dataset by $99.8\%$ from $2.37 \pm 0.0279$ seconds when using the $FOT$ planner-based evaluator to $0.00572 \pm \num{3.50e-5}$ seconds, a $414.34\times$ latency improvement, per time step per actor per goal. On the simulated dataset, the NN-based approximator reduces the reachability compute latency from $2.087 \pm 0.0030$ seconds when using the $FOT$ planner-based evaluator to $0.00569 \pm \num{2.15e-5}$ seconds, a $366.7\times$ latency improvement.

\subsection{Efficacy of RL-based safety mitigation controller}
\label{s:smc_results}
\begin{figure}[t!]
	\centering
	\subcaptionbox{~\label{subfig:ghostcutin_prob}}{\includegraphics[width=0.23\textwidth]{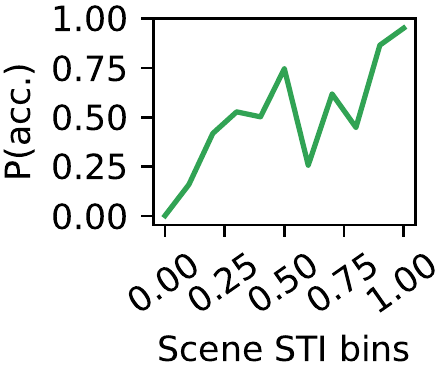}}
	\subcaptionbox{~\label{subfig:miti_all_ghost}}{\includegraphics[width=0.23\textwidth]{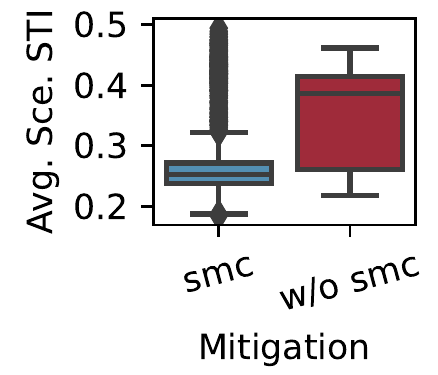}}
	\caption{\label{fig:miti_plot} (a) Accident probability $P(acc.)$ vs \score binned values of the cut-in scenario mutations, (b) Mean scene \score distribution with and without SMC of the cut-in scenario mutations }
\end{figure}

The strong correlation between the \score value and the probability of an accident suggests that \score value is a good indicator of an imminent accident. However, instead of mitigating only when the \score is high (reactive mitigation), it is more effective to proactively reduce the \score as outlined in \cref{s:rl_agent1} to avoid high \score situations. 
\cref{tab:miti_results} shows the accident rate comparison between the ego actor (the LBC agent) with SMC, and the ego actor without SMC for the variant scenarios. 
The SMC reduces the accident rate from $0.565$ to $0.128$, a $77.3\%$ reduction, on the cut-in scenario mutations.
This is reflected in the higher mean \score values of the variant scenario runs in which the ego actor without SMC leads to accidents, shown in \cref{subfig:miti_all_ghost}. Across the 1000 cut-in scenario mutation executions, the SMC reduces the scenario \score, which leads to a lower accident rate. 
Refer to Supplementary Material for illustrative examples of SMC-based mitigation.

The high mitigation efficacy in all variant scenarios suggests that the SMC can generalize to variant scenarios with similar topology but different hyperparameters. It achieves this efficacy by learning from just a few instances of the variant scenarios of that typology.  In our case, the SMC is trained on one of the variant scenarios and evaluated on all variant scenarios (recall \cref{s:train_val}).

\section{Conclusion and Future Work}
\label{ml4ad:s:conclusion}
We proposed a novel safety-threat indicator (\score) that allows us to identify risky actors and driving scenarios. 
Using \score as a basis, we developed a safety-hazard mitigation controller (SMC) to maximize safety for an AV by focusing on the actors most threatening to the AV's safety. 
The SMC's
execution is independent of the existing AV controller, so it does not interfere with normal AV operations.
Our future work will address limitations and explore further use-cases of \score for other autonomous driving tasks.

Three areas remain to be addressed. First, the proposed indicator (and hence the controller) assumes that each navigational choice is equally safe.
However in practice, certain choices are safer than others. For example, in our grid formulation, we can estimate time-to-collision in each grid to further rank each path (i.e., each trajectory from the current position to one of the grid locations). Second, the SMC was only trained and evaluated in simulation. Given the hazardous nature of these scenarios, there is a need to evaluate the robustness of the proposed controller in a real-world, on-road environment and identify techniques to transfer the model from simulation to the real world.
And third, we need to evaluate the sensitivity of \score to different planners for evaluating reachability (e.g., R2P2~\cite{Rhinehart_2018_ECCV} and  RRT~\cite{lavalle1998rapidly}). 
In the future, we will both address these limitations and apply \score in other use-cases, including the following: \score can assist with rare test dataset/benchmark generation for autonomous driving tasks. For example, \score can be integrated with fuzzing and adversarial techniques to identify cases in which AVs will collide~\cite{9564360,9251068}. 
\score can also be directly integrated into the planner~\cite{ fisac2019hierarchical}, alleviating the need for the mitigation controller.
And finally, \score can be integrated with systems like PyLot~\cite{pylot} and others~\cite{li2020towards} for deadline-aware scheduling tasks.

{
\bibliographystyle{ieeetr}
\bibliography{references}
}

\newpage
\appendix
\section*{Appendix}

\begin{table}[h]
	\centering
	\begin{tabular}[t]{c|l}
		\toprule
		\textit{Mitigation policy}  & \makecell{\# acc./\# T.runs} \\  %
		\midrule                             
		\textit{\small{LBC}}                              &  565/1000 (0.565)   \\ \hline %
		\textit{\small{RIP (WCM)}}                              &  478/1000 (0.478)   \\ \hline %
		\makecell{\small{\textit{LBC+SMC}} (ours)} &  \textbf{128/1000 (0.128)}   \\ %
		\bottomrule
	\end{tabular}
	\vspace{+2mm}
	\caption{\label{sm_tab:addi_res} Comparison of mitigation policies (i) \textit{LBC} agent \cite{chen2019lbc}, (ii) \textit{RIP} agent with the worst case model (WCM) \cite{pmlr-v119-filos20a}, and (iii) \textit{LBC} agent with SMC (ours); \# acc.: number of of runs with accident, \#T.runs: total number of runs.}
\end{table}
\section{Additional Experiments and Results}
\subsection{RIP agent as an additional baseline for SMC evaluation}
In the main paper we compare the SMC's mitigation efficacy with the baseline agent-- the \textit{LBC} agent by Chen et. al.~\cite{chen2019lbc}. 
Here we add the \textit{RIP} agent by Filos et. al.~\cite{pmlr-v119-filos20a} as an additional baseline ego actor for comparison. 
Since authors only provides the training dataset and training code but not the model weights, we train the \textit{RIP} agent on
the provided training dataset using 4 ensemble models. Note that the
authors do not provide the number of epochs used for training each
model, so we train each model until the loss converges.
We compare our SMC with the \textit{RIP-worst case model} or RIP-WCM as it is the best performing model in \cite{pmlr-v119-filos20a}.

Similar to the \textit{LBC} agent, we evaluate the \textit{RIP} agent on the $1000$ mutated scenarios with the same simulation environment and hardware and report its accident rate in \cref{sm_tab:addi_res}. Though out performing the \textit{LBC} agent, the \textit{RIP} agent results in a high accident rate of 0.478 comparing to the \textit{LBC+SMC} (our) approach which results in an accident rate of 0.128 on the same mutated scenarios. Our approach reduces the accident rate by 73.2\% compared to the \textit{RIP} agent.

The \textit{RIP} agent improves safety by leveraging the fact that out of distribution data (OOD) leads to higher epistemic uncertainty, and hence the agent should act conservatively. 
In the WCM ensemble setting, the \textit{RIP} agent acts according to the most permissive plan and this can lead to recovery from out of data distribution. 
There is a possibility that the detection of the distribution shift is not done sufficiently in advance and, therefore, there might not be any safe trajectory available to the ego agent at the time of detection.
In other words, even the most permissive plan is not safe; thereby, leading to accident as evident from our evaluation.
Instead, the SMC learns the mitigation actions by embedding the knowledge of \score--an indication of safety threatening situation--as part of its policy which enable it to mitigate dangerous situations (recover from OOD) more effectively. 
Moreover, we assert that the SMC approach is orthogonal to approaches like \textit{RIP} or \textit{LBC} and they can be applied simultaneously to improve safety and robustness of autonomous driving.

\subsection{SMC mitigation action illustration with an example}
\begin{figure}[h]
	\centering
	\subcaptionbox{~\label{subfig:nomiti_run_sti}}{\includegraphics[width=0.25\textwidth]{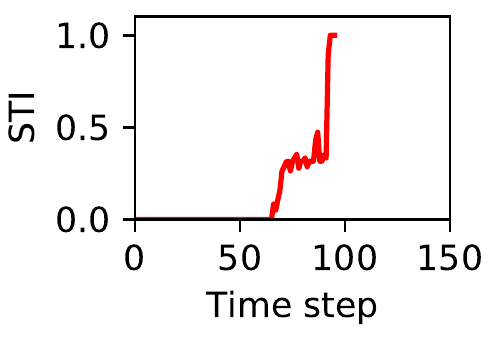}}
	\subcaptionbox{~\label{subfig:miti_run_sti}}{\includegraphics[width=0.25\textwidth]{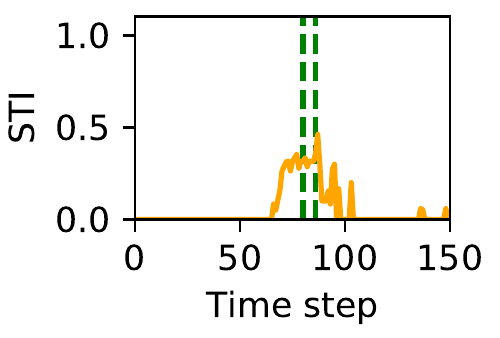}}
	\subcaptionbox{~\label{subfig:miti_scenes}}{\includegraphics[width=0.251\textwidth]{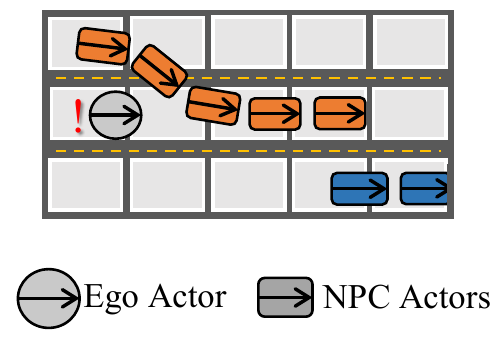}}
	\caption{\label{fig:miti_run} Scene \score time series of a mutated scenario run. (a) shows scene \score in red without SMC enabled. (b) shows scene \score with SMC enabled and mitigation period between dashed lines. (c) Shows the moment of cut in typology.}
\end{figure}
Here we show an example activation of the SMC's mitigation action.
SMC proactively applies emergency braking proactively to reduce the \score in a safety critical scenario and, hence, avoids the accident which otherwise would happen (using the \textit{LBC} as the ego actor). 
\cref{fig:miti_run} plots the scene \score of the scenario per time step.
In \cref{subfig:miti_run_sti}, the orange line represents the running scene \score with the SMC enabled and the green dashed vertical lines shows the start and the end of the mitigation action period. The red line in \cref{subfig:nomiti_run_sti} represents the running \score without SMC for comparison. 

In this run, as the threatening NPC actor approaching from behind the ego actor at a higher speed, the driving flexibility decreases (\score increases) at the $60^{th}$ time step. At around the $80^{th}$ time step right before the cut in happens, the \score starts to increase (in both \cref{subfig:miti_run_sti,subfig:nomiti_run_sti}) and subsequently the SMC intervenes and brakes in \cref{subfig:miti_run_sti} (after the first green vertical line) to reduce the \score and avoid possible accident, and SMC then disengages as the situation alleviates (after the second green vertical line) and the ego actor continues to drive to its destination. Without the SMC, the ego actor alone \cref{subfig:nomiti_run_sti} (using only the LBC agent) fails to recognize this adverse situation and misses the mitigation window, resulting in an accident. We can see in \cref{subfig:nomiti_run_sti} the \score continues to increase (i.e., red line continues to climb) until the time of the accident ($100^{th}$ time step) after which the simulation terminates.

\subsection{High \score scenes from real-world dataset}
\begin{figure}[h]
	\centering
	\includegraphics[width=0.95\textwidth]{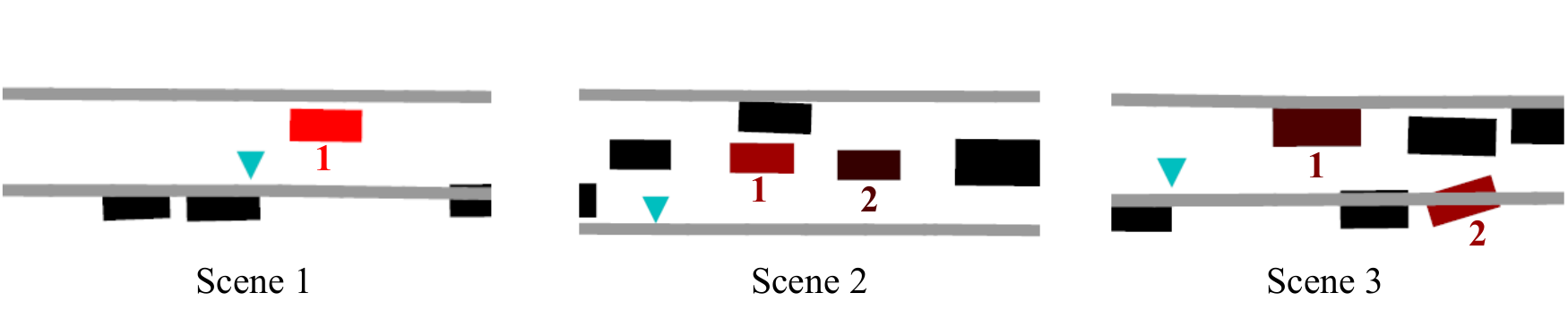}
	\caption{\label{fig:hi_sti_scene} BEV representation of real-world high \score scenes from Arogverse dataset. Here, the blue triangle represents the ego actor and the rectangles represents NPC actors. The redness of an NPC actor indicates the \score value--the brighter the red color, the higher the \score of that NPC relative to other NPC actors' \score in that scene. Labeled NPC actors with non-zero \score are labeled.}
\end{figure}
Recall from \cref{s:real_wor_res} that we evaluate the actors' and the scenes' \score on the \textit{Argoverse} real-world dataset.
The result show that high \score actors or scenes are rare because real-world datasets are collected in a safe environment with human supervision. 
We extract those rare scenarios and actors using our proposed indicator.

We find that a scene \score can be high even with low \score actors in safety-critical configurations.
Here we show examples of high \score scenes, as BEV representations, from the \textit{Argoverse} dataset and provide the reasons why the scenes' \score are high based on the actors configuration and the actors' \score by exploiting the interpretability of our formulation.

\paragraph{Scene 1: High scene \score due to one actor.} In the first scene (\cref{fig:hi_sti_scene} Scene 1), the ego actor travels on the lane with NPC actor 1 traveling closely in the adjacent lane. The scene \score of this scene is $0.643$. The actor \score for NPC actor 1 is $0.642$, with other actors having 0 \score. In this case, the NPC actor 1's \score is high because it travels closely with the ego actor which reduces the ego actor's driving flexibility considerably. Moreover, the NPC actor 1 is solely responsible for the high scene \score because the ego actor's driving flexibility is constraint mainly by NPC actor 1, while other NPC actors have limited affect on the ego actor's driving flexibility.  

\paragraph{Scene 2: High scene \score due to the presence of NPC actors in the close vicinity.} In the second scene (\cref{fig:hi_sti_scene} Scene 2), the ego actor travels on a multi-lane busy highway.
NPC actor 1 and 2 are traveling in the same direction on the left adjacent lane in close vicinity.
The scene \score of this scene is high at $0.625$. However, the actor \score for the NPC actor 1 and 2 are relatively low compared to the scene \score at $0.19$ and $0.063$ respectively, with other NPC actors having 0 \score.
Note that the sum of individual actor \score does not need to be equal to the scene \score because the reduction in the ego actor's driving flexibility is jointly constrained by the individual actor's trajectories.
 In this case, the scene \score is high because the driving flexibility is greatly constraint by NPC actors 1 and 2 jointly and the ego actor can only stay in its own lane. 
 Removing all NPC actors in the scene greatly improves driving flexibility as the ego actor can now chose to merge left. On the other hand, the \score of each individual NPC actor (NPC actor 1 and 2) is low because each NPC actor only contributes partially to the reduction of the ego actor's driving flexibility so removing only one NPC actor only increase the driving flexibility marginally. 
 For example, removing NPC actor 1 only frees up the adjacent space near to the ego actor, the ego actor is still constraint by NPC actor 2.
Finally we can see that the actor's \score distribution is sensitive to distance which a higher \score is assigned to the closer actor (NPC actor 1) as it limits the driving flexibility more than NPC actor 2 that is farther ahead. 

\paragraph{Scene 3: High scene \score due to the intrusive driving maneuver of NPC actors.} In the third scene (\cref{fig:hi_sti_scene} Scene 3), the ego actor travels on the right lane going straight with some NPC actors stopped on the left lane waiting to turn left and some NPC actors park on the right. 
In particular, NPC actor 1 stops on the left lane waiting to turn left, and NPC actor 2 starts merging from the parking lane into the ego lane.
The scene \score is high at $0.64$, while NPC actor 1 and NPC actor 2 have \score of $0.09$ and $0.18$, receptively. As Scene 2 described above, the scene \score is high because the ego
actor's driving flexibility is constraint by the NPC actors and in this case collectively and the ego actor can only choose to merge left and stop or continue on the same lane. On the other hand, the individual NPC actor's \score are low because removing each of the NPC actor (NPC actor 1 or 2) only improves the driving flexibility marginally, with other NPC actors still constraining the ego actor's driving flexibility. 
However, Different from Scene 2, the actor's \score distribution in this scene is sensitive to motion as the NPC actor 2, the actor that starts driving and merging in front of the ego actor, is assigned a higher \score value than the NPC actor 1, which stops for traffic light, despite NPC actor 1 is closer to the ego actor.

\section{\score Formulation in the Realtime Setting}
This section outlines the detailed derivation of \score evaluation in the realtime setting as a supplementary to \cref{s:sim_rt}. Recall that in the oracle setting we assume that the set of trajectories of all actors from $t=0$ to $t=T$, $\mathbb{X}_{t:t+k}$, is precisely known (without uncertainty). This assumption does not hold in the realtime setting in which the actors' past, current, and future states by nature are noisy as they are estimated from realtime sensor measurements. Hence, we extend our formulation of the oracle setting to account for the actors' noisy state estimations and trajectory set, as follows.

\paragraph{Obtaining $\bar{X}^{(i)}_{t:t+k}$ for an actor.} Recall that $\bar{X}^{(i)}_{t:t+k}$ is the set of states from $t$ to $t+k$ of actor $i$. 
In the realtime setting, the ego actor uses the past and the current state measurements $o_{0:t}^{(i)}$ to estimate the current state $\bar{x}^{(i)}_t$ of an actor. Note that $\bar{x}^{(i)}_t$ is different from $x^{(i)}_{t}$ (defined in the oracle setting in \cref{s:sim_gt}), as $\bar{x}^{(i)}_t$ captures the uncertainty/noise associated with the state.
The horizontal bar on the top of the symbols is used here to distinguish between the noisy state data and ground truth state data. 
Typically,  $o_t^{(i)}$ is the bounding box that is detected using object-detection algorithms, such as YOLO~\cite{redmon2018yolov3}, and $\bar{x}^{(i)}_t$ is estimate using a state-space estimation algorithm $f_{esitmate}$, such as \textit{Kalman Filter} \cite{welch1995introduction} from measurement history $o_{0:t}^{(i)}$. The future states of an actor $i$ from time $t$ to time $t+k$ can then be predicted given a state prediction model $f_{predict}$ and the actor's state estimation history $\bar{x}^{(i)}_{0:t}$:
\begin{equation}
	\bar{X}^{(i)}_{t:t+k} = f_{predict}(\bar{x}^{(i)}_0, ..., \bar{x}^{(i)}_{t}) = f_{predict}\left(f_{esitmate}(o^{(1)}_{0}, ...,  o^{(N)}_{t})\right)
	\label{eq:s42}
\end{equation}
Alternatively, given that state estimations and predictions often follow the algorithm's underlying noise assumptions \cite{welch1995introduction}, $\bar{X}^{(i)}_{t:t+k}$ can also be modeled by the probability distribution shown in \cref{eq:s43}
\begin{equation}
	\bar{X}^{(i)}_{t:t+k}= P(\bar{x}^{(i)}_t, ..., \bar{x}^{(i)}_{t+k}| o^{(1)}_{0}, ...,  o^{(N)}_{t})
	\label{eq:s43}
\end{equation}

\paragraph{Constructing $\mathbb{Q}_{t:t+k}$ for \score evaluation.} Recall from above that $\bar{X}^{(i)}_{t:t+k}$ is the future state estimation distributions due to the noisy state estimation of an actor $i$, from time $t$ to $t+k$. Let $\mathbb{Q}_{t:t+k}$ denote the set consisting of samples of future trajectories for all actors, we can construct $\mathbb{Q}_{t:t+k}$ by sampling from $\bar{X}^{(i)}_{t:t+k}$ for each actor $i$ for all $i$: 
\begin{equation}
	\mathbb{Q}_{t:t+k} = \{ {Q^{(1)}_{t:t+k} \sim {\bar{X}}^{(1)}_{t:t+k}, ..., Q^{(N)}_{t:t+k}} \sim {\bar{X}}^{(N)}_{t:t+k} \}
	\label{eq:s44}
\end{equation}
Following the oracle setting, let us assume that an oracle planner $f_{planner}$ exists and can generate all possible future trajectories that the ego actor can follow safely given $\mathbb{Q}_{t:t+k}$, $\mathbb{M}$, and ${x}_{t}^{ego}$ (\cref{eq:s45}).
\begin{equation}
	\bar{Z}_{t:t+k} = f_{planner}(\mathbb{M}, \mathbb{Q}_{t:t+k}, {x}^{ego}_{t})
	\label{eq:s45}
\end{equation}
Then we can evaluate the actor \score ($\rho^{(i)}_{t:t+k}$) and scene \score ($\rho_{t:t+k}$) with equation \cref{ml4ad:eq:actorrisk} and \cref{ml4ad:eq:totalrisk}, respectively with the sampled trajectory set $\mathbb{Q}_{t:t+k}$.

Note that, for different sampled instances of $\mathbb{Q}_{t:t+k}$ with \cref{eq:s44} $\bar{Z}_{t:t+k}$ will be different. Therefore, we can estimate the uncertainty in the \score of an actor and the scene by sampling multiple instances of $\mathbb{Q}_{t:t+k}$ and executing \cref{ml4ad:eq:actorrisk} and \cref{ml4ad:eq:totalrisk} multiple times.

\section{BEV Representation of Driving Scenes}
\label{sm:bev_rep}
In order to compute \score based on reachability using commonly employed planners $\hat{f}_{planner}$ such as RRT~\cite{lavalle1998rapidly}, FOT~\cite{5509799}, and Hybrid A*~\cite{dolgov2008practical}, we need to represent the driving scene in planner recognizable formats. Here we show the construction of the BEV representation for the planner to consume in details.

\paragraph{Representing $\mathbb{X}_{t:t+k}$ and $x_t^{ego}$.} Referring to \cref{eq:reach_planner}, one part of our representation of the driving scene per time step $t$ includes the NPC actors' projected future trajectories $\mathbb{X}_{t:t+k}$, and the current ego actor state $x_t^{ego}$. Collecting information for $\mathbb{X}_{t:t+k}$ and $x_t^{ego}$ is straight forward as these are either provided by the dataset's ground truth labels or the state of a simulator (as ground truth labels) in the oracle setting, or by estimating them from the sensor measurements in the real-time setting. The state of an actor consists of the center location and the size of its bounding box. The state of the ego vehicle consists of the location, rotation, and the velocity.

\paragraph{Representing $\mathbb{M}$ and $G_k$.} The other part of our representation includes representing the map information $\mathbb{M}$ and the goal information $G_k$, as shown in \cref{eq:reach_planner}, in a planner consumable format. We represent $\mathbb{M}$ by dividing the map space into "drivable" and "non-drivable" areas. Drivable ares are road surfaces that belongs to a drivable lane, which includes the current ego lane (the lane that the ego actor is traveling on) and adjacent lanes traveling in the same direction. 
Non-drivable areas are road surfaces that are either non-drivable (e.g., passer-by or parking lots), or belong to lanes traveling in the other direction. For example, given a two way street, the lanes traveling in the same direction as the ego lane are drivable areas, and everywhere else are non-drivable areas. To constraint the planner to consider drivable path only, we further construct $B$ as a set of static obstacles from the lane boundaries.

We further divide drivable areas into a set of grids, denoted $C_k$. Each grid is of size $l \times w$ with the center of the grid being a potential goal location $g_k$, where $l$ is the length of the grid in the longitudinal direction of the lane, and $w$ is the width of the grid in the lateral direction of the lane. The choice of $l$ and $w$ can be flexible, for example, in our case for the \textit{Argoverse} dataset which the length width is not specific, we choose $l = 5$ meters and $w = 3.7$ meters, corresponding to the common length of a family car and the common width of a motorway lane. The simulator dataset provides the lane width information to set $w$.
Note that given the current ego actor state $x_t^{ego}$, its dynamic constraints, and the time budget $k$, the farthest distance $d$ that the ego actor can reach is governed by
\begin{equation}
	d = |\mathbf{v}_{t}^{ego}| + k * |\mathbf{a}_{max}^{ego}|
	\label{sm_eq:d}
\end{equation}
given the lane is not infinitely wide, the number of grids that the ego actor can potentially reach is also finite. Moreover, it is a reasonable assumption that the ego actor should not plan to travel backward, so the set of potential goals $G_k$ given the time budget $k$ can be expressed as 
\begin{equation}
	G_k = \{g_k | g_k \text{ is of forward direction of ego actor } \cap dist(g_k, x_{t}^{ego}) \le d, g_k \in C_k\}
\end{equation}

\paragraph{Representation in ego coordinates.} For convenience, we transform the actor trajectories $\mathbb{X}_{t:t+k}$, the ego actor state $x_t^{ego}$, the set of potential goals $G_k$, and the set of static lane boundaries $B$ to ego-centered coordinates. That is, we first transform the locations of all the aforementioned sets by subtracting the ego actor's location, and then we rotate them in the opposite direction of the ego actor's rotation about the origin, per time step $t$. We can then drop the location and rotation element from the ego state since they are always zeros in the ego-centered coordinate system. 

Given that $(G_k, \mathbb{X}_{t:t+k}, x_t^{ego}, B)$ is the BEV representation of the scene at time step $t$, we can now express \cref{eq:reach_planner} in the new terms to calculate the set ${R}_k$ using $\hat{f}_{planner}$
\begin{equation}
	{R}_k = \{g_k | g_k \in G_k, \mathbbm{1}_{reachable}\left[\hat{f}_{planner}(g_k, \mathbb{X}_{t:t+k}, x_t^{ego}, B, k)\right] = 1\}
	\label{eq:reach_planner}
\end{equation}
\begin{figure}[t!]
	\centering
	\subcaptionbox{~\label{subfig:bev_real1}}{\includegraphics[width=0.45\textwidth]{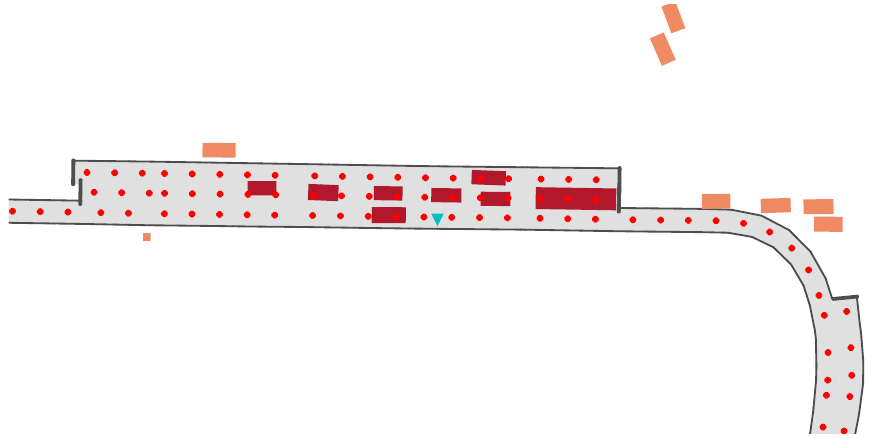}}
	\subcaptionbox{~\label{subfig:bev_real3}}{\includegraphics[width=0.45\textwidth]{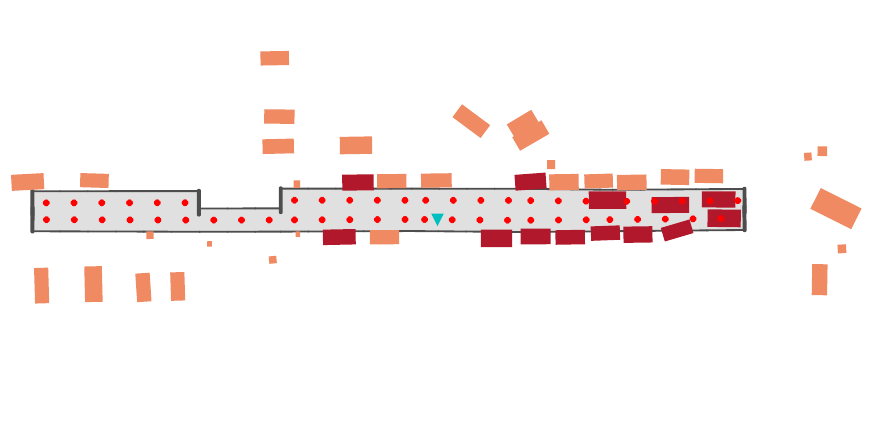}}
	\subcaptionbox{~\label{subfig:bev_real3}}{\includegraphics[width=0.45\textwidth]{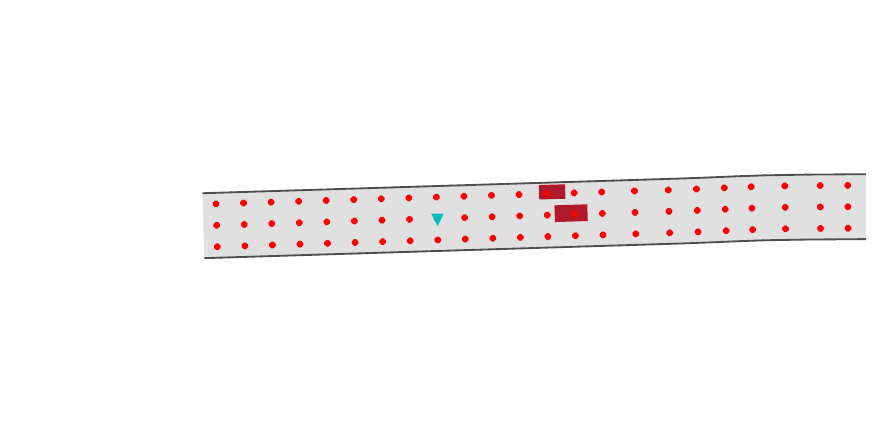}}
	\subcaptionbox{~\label{subfig:bev_real3}}{\includegraphics[width=0.45\textwidth]{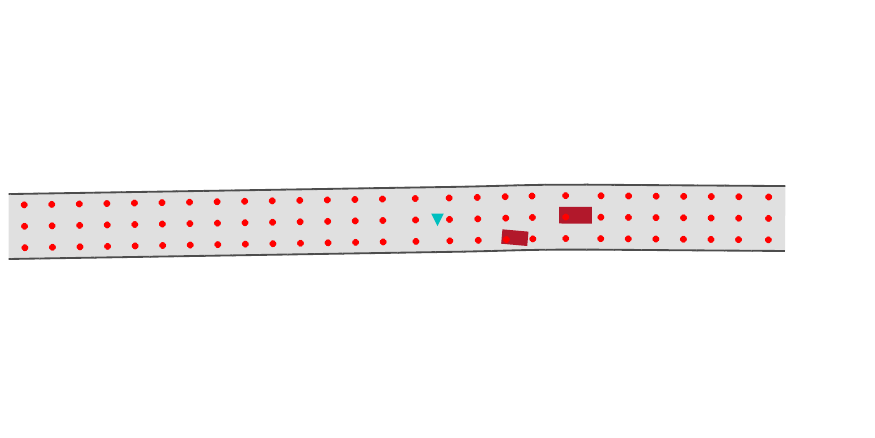}}
	\caption{\label{fig:bev_real} BEV representation of scenes from the \textit{Argoverse} dataset are shown in subfigure (a) and subfigure (b). Similar representation of scenes from the \textit{Carla} simulated dataset are shown in subfigure (c) and subfigure (d). In all subfigures, the red dots represent the center of a grid and potential local goals, gray surfaces represent drivable lane surfaces, gray boundaries represent lane boundaries, red rectangles represent on lane NPC actors, orange rectangles represent off lane NPC actors, and the blue triangle represents the ego actor.}
\end{figure}
\paragraph{BEV Visualization.} \cref{fig:bev_real} shows visualizations of the BEV representation of scenes from the \textit{Argoverse} dataset and the \textit{Carla} simulator. The Drivable area is shown in gray with potential \textit{local goals} shown as red dot. The Lane boundaries are shown in dark gray around the drivable area as static obstacles $B$. The actors that are on-lane are shown in red, and actors that are off-lane are shown in orange. Finally, since the BEV representation uses a ego-centered coordinate system so it is always the middle of the figure, shown in light blue triangle.

\section{Illustration of \score calculation using reachability.}
\begin{figure}[t!]
	\centering
	\includegraphics[width=0.99\textwidth]{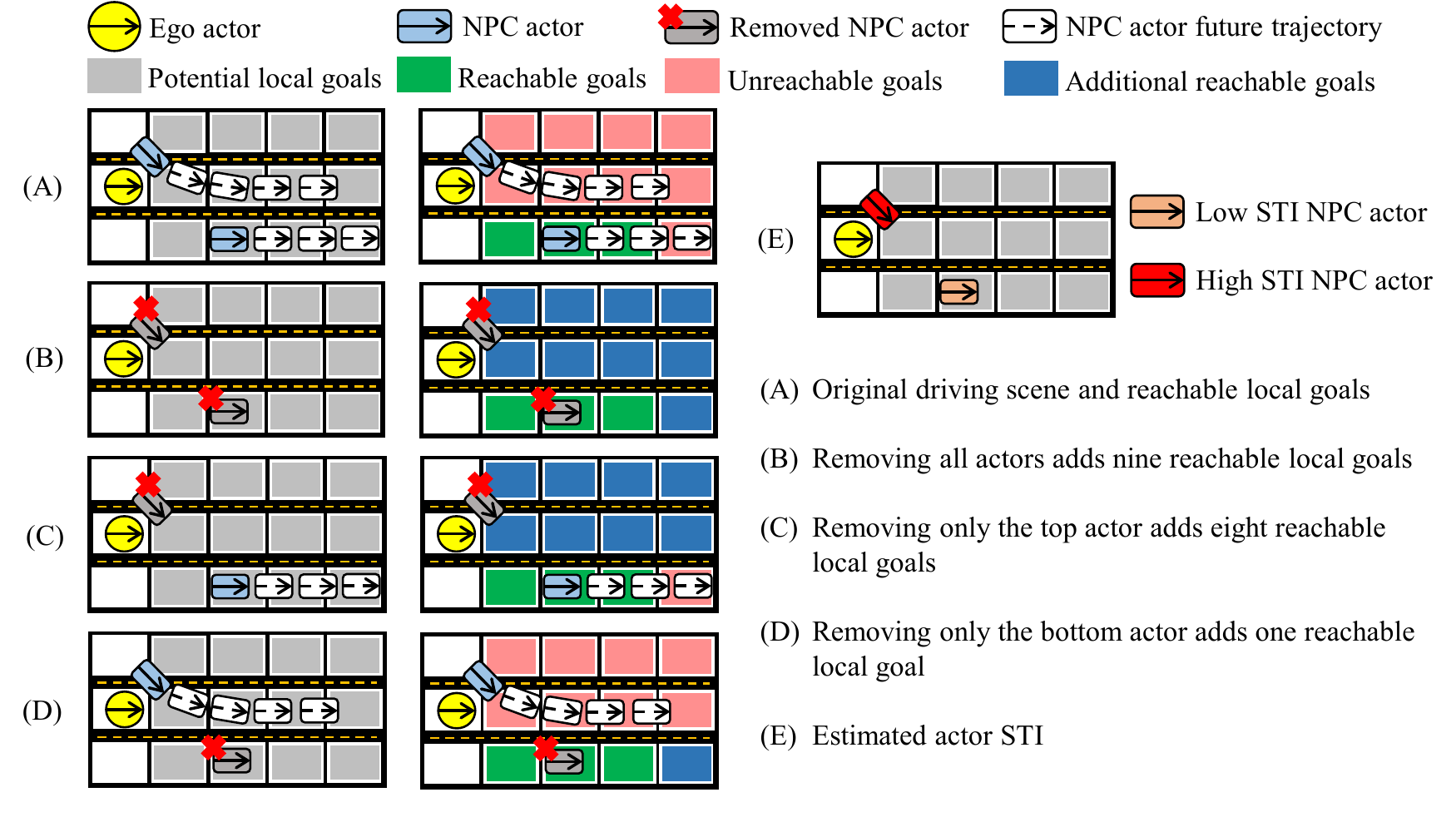}
	\caption{\label{fig:reach_cal} \score calculation using discretized drivable areas and reachability to the grids. Subfigure (A) shows \score calculation with all actors present, (B) shows \score calculation with all actors removed, (C) or (D) shows \score calculation with one of the actor removed, and (E) shows the relative actor \score in this scene.}
\end{figure}
\cref{fig:reach_cal} illustrates the scene and actor \score calculation via discretized drivable areas and reachability (recall \cref{ml4ad:s:design} in the main paper) in a two NPC actors scene, in which one NPC actor cuts in front of the ego actor from the left and the other NPC actor travels straight in the right lane. 

Starting with the original configuration as shown in \cref{fig:reach_cal} (a), we first calculate the local goals' reachability with all NPC actors present. After accounting for the NPC actors' trajectories into the future and the ego actor's dynamic capability, only three cells are reachable. Note that even though it appears that the bottom NPC actor occupies certain local goals in the bottom lane, those goals are in fact reachable by the ego actor as the bottom NPC actor will move ahead in the next $k$ (recall $k$ is the time budget) time steps. 

Then, (B) calculates the reachability to the local goals with all NPC actors removed. After removing all NPC actors, all local goals are reachable by the ego actor. Compared to (A) the original setting, removing all actors increases driving flexibility by nine local goals. After reachability calculation of (A) and (B) we have sufficient information to calculate the scene \score using \cref{eq:reach_totalrisk}. 

To calculate \score for a particular actor $i$ via counterfactual reasoning actor $i$ is remove from the scene and the reachability is recalculated. Together with the reachability information on scene (A) and (B), we can then calculate the \score for $i$ using \cref{eq:reach_actorrisk}. This is shown in (C) and (D) in \cref{fig:reach_cal}, with (C) removes the top actor, and (D) removes the bottom actor. Because the top actor resulted in a larger driving flexibility reduction--removing it increases driving flexibility by eight goals compared to one goal by removing the bottom actor---the top NPC actor's \score is higher as illustrated in (E) in which only the current location of the actor is shown. 

\section{NN-based Approximator Additional Details}
\subsection{Dataset generation}
\label{sm:data_gen}
The input to the neural network based reachability approximator is a set of stacked $k$ bird-eye-view (BEV) images $\mathbf{I}_{t:t+k}$ and the vector $\mathbf{e}_t$ that contains the ego actor's current state (see \cref{s:nn_reach}). $\mathbf{I}_{t:t+k} = \{I_t, I_{t+1},...,I_{t+k}\}$ can be constructed from the BEV representation $(G_k, \mathbb{X}_{t:t+k}, x_t^{ego}, B)$, where $G_k$ is the set of potential goals, $\mathbb{X}_{t:t+k}$ is the set of actor states for all actors from $t$ to $t+k$, $x_t^{ego}$ is the current state of the ego actor, and $B$ is the set of static road boundaries (treated as static obstacles).

Let's first take time step $t$ and generation of the BEV image $I_t$ as an example. Recall that the BEV representation $(G_k, \mathbb{X}_{t:t+k}, x_t^{ego}, B)$ uses the ego-centered coordinate system and the goal is to convert such a representation in to a BEV binary image in which the drivable areas, such as open lane space, are labeled as 1, and the undrivable areas, such as those that are occupied by the NPC actors or out-of-lane, are labeled as 0. 

\paragraph{Generating BEV binary images $\mathbf{I}_{t:t+k}$.} We first convert the ego-centered coordinate into image-coordinates, in terms of pixels. Following the convention of a 2D array representation, we set the direction from top to bottom as the positive x-direction, and from left to right as the positive y-direction, the origin of the image is at the top-left corner. Suppose we limit the view ranges of the x-direction to $[-R_X, +R_X]$ and the y-direction to $[-R_Y, +R_Y]$ in the ego-centered coordinates BEV representation, we can then generate a top-down abstract BEV visualization (it is abstact because the coordinates are continuous) of the BEV representation in terms of ego-centered coordinates, with the ego actor at the center and the positive y-direction (to the right of the visualization) being the forward direction. The visualization is only generated for the range between $[-R_X, +R_X]$, and $[-R_Y, +R_Y]$ and cuts off otherwise. In this visualization, areas that are part of the drivable lanes (i.e.,are not occupied by $\mathbb{X}_{t}$ or $B$) can be labeled as 1, and the rest 0. Suppose we denote this abstract visualization as $V_t$ then for each location $(X, Y) \in [-R_X, +R_X] \times [-R_Y, +R_Y]$
\begin{equation}
	V_t(X, Y) = 1 \text{ if } (X, Y) \notin \mathbb{X}_t \land (X, Y) \notin B, \text{ else } V_t(X, Y) = 0
\end{equation}
We can then convert this abstract BEV visualization $V_t$ into a binary image $I_t$ by applying the transformation that maps ego-centered coordinates to image pixel coordinates. Suppose the input image $I_t$ is of size $H\times W$ then each pixel represents a distance in the x-direction of $\frac{2R_X}{H}$ meters, and distance in the y-direction of $\frac{2R_Y}{W}$ meters. The transform that maps ego-centered coordinates $(X,Y)$ to image pixel coordinates $(x,y)$ is defined as follows:
\begin{equation}
	(x, y) = \left( \left\lfloor {\frac{X+R_X}{2R_X/H}} \right\rfloor, \left\lfloor {\frac{Y+R_Y}{2 R_Y/W}} \right\rfloor \right)
\end{equation}
To construct the BEV image for the future time steps, i.e., $I_{t+1}$ to $I_{t+k}$, we can simply replace the set of current actor states  $\mathbb{X}_t$ with the state predictions for future time steps: $\mathbb{X}_{t+1}$, $\mathbb{X}_{t+2}$ etc., while other states the same. The $\mathbf{I}_{t:t+k} = \{I_t, I_{t+1},...,I_{t+k}\}$ can then be constructed from the set of BEV images just generated.
\paragraph{Labeling reachability indicator vector $\mathbf{r}_{k}^*$.} The output of the NN-based reachability approximator is (see \cref{ml4ad:eq:nn_planner}) is a reachability indicator vector $\mathbf{r}_{k}$ which each element indicating a location on the BEV image as reachable or not reachable. Since the neural network's output size is fixed, we choose to discretize the entire BEV image into grids instead of using the original discretized grids of drivable areas in which the number of grids may vary. Below we describe the process of generating the ground truth label $\mathbf{r}_{k}^*$ for training and evaluation.

Suppose each grid is of size $a \times b$ pixels in the BEV image space. A grid of size $a \times b$ in the image space is equivalent of a grid of size $\frac{2bR_Y}{W}$ meters in the y-direction, and $\frac{2aR_X}{H}$ meters in the x-direction in the ego-centered coordinate system (used by $V_t$). Given a BEV image size of $H \times W$, there are in total $\left\lfloor \frac{H}{a} \right\rfloor \times \left\lfloor \frac{W}{b} \right\rfloor$ grids. Notice that goals are only in the forward direction as we assume the ego actor cannot travel backward, so we can remove the grids for the left half of the BEV image because they contain no goals. $\mathbf{r}_{k}^*$ can then be constructed from a vector of all zeros of size $\left\lfloor \frac{H}{a} \right\rfloor \times \left\lfloor \frac{W}{2b} \right\rfloor$, then setting the elements which overlaps with a reachable goal $g_k \in R_k$ to 1, where $R_k$ is the output of \cref{eq:reach_planner}:
\begin{equation}
	\mathbf{r}_{k}^*[i] = 1 \text{ if } i^{th} \text{ grid contains a reachable goal } g_k \in R_k \text{ in the image space, else 0.} 
\end{equation}
We choose $H=128$px, $W=512$px, $R_X = 17.5$m, $R_Y = 70$m, $a = 11$px, $b = 20$px for all the experiments. The training dataset is generated from \textit{Argoverse} training dataset which consists of $\sim 7800$ datapoints. The testing dataset is generated from the \textit{Argoverse} validation dataset which consists of $\sim 930$ datapoints. Furthermore, we perform a 80\%-20\% split on the training data set for train-validation during the training procedure.

\paragraph{Dataset visualization.}
\begin{figure}[t!]
	\centering
	\includegraphics[width=0.80\textwidth]{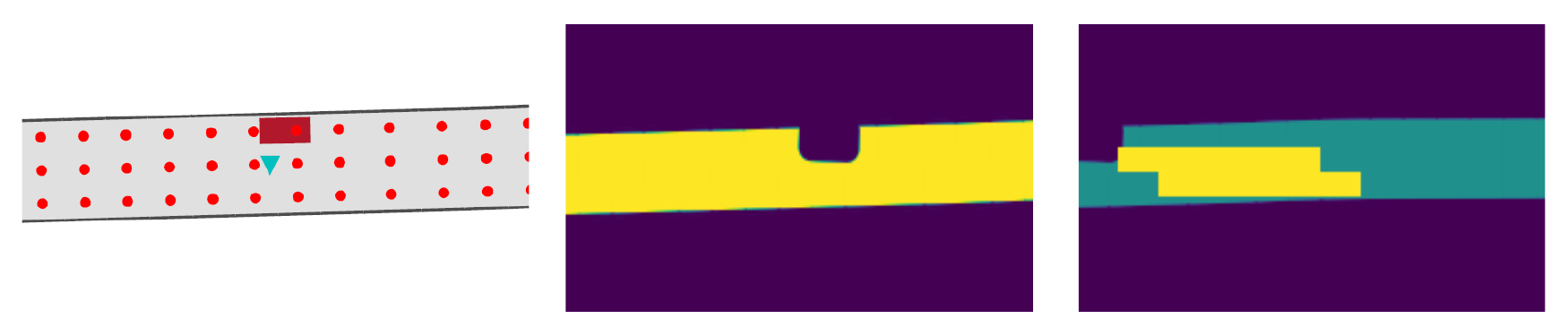}
	\caption{\label{fig:data_vis} NN-based reachability approximator training and evaluation dataset visualization. The left subfigure shows the BEV representation of a scene. The middle subfigure shows a part of the input data $I_t$ based on the BEV representation. The right subfigure visualizes the label $\mathbf{r}_t$. In the label visualization (the right subfigure), dark areas are undrivable ares, greenish-blue areas are drivable lanes, and the yellow grids represents reachable grids (elements of $\mathbf{r}_t$ whose values are 1) in the image space.}
\end{figure}
\cref{fig:data_vis} shows the visualization of the generated dataset. The left subfigure shows the original BEV representation of the scene, with the red-dots being the potential local goals, red blocks being an NPC actor, and drivable areas in gray. 
The middle subfigure shows the binary BEV image $I_t \in \mathbf{I}_{t:t+k}$ as part of the input to the model, with bright areas being the drivable areas and dark areas being the undrivable areas; notice that the NPC actor is part of the undrivable area. 
The right subfigure visualizes the drivable (in greenish-blue), non-drivable areas (dark), and reachable grids indicated by $\mathbf{r}_k$ (yellow). 
The drivable areas that are not reachable by the ego actor due to the presence of the NPC actor aren't highlighted in yellow.

\subsection{NN-based reachability approximator architecture}
\begin{table}[]
		\begin{tabular}[t]{l|l|l}
			\toprule
			\textit{Layer(s)} & \makecell{\textbf{Input dim}}  &  \makecell{\textbf{Output dim}} \\ 
			\midrule
			\textit{Fully connected + ReLU}         & $\mathbf{e}_t$: $(1 \times 5)$                &  $(1, H, W)$   \\  \hline
			\textit{Concatenate [$\mathbf{I}_{t:t+k}$, $\mathbf{e}_t$]}             & $Concat([ (k+1, H, W)$, $(1, H, W)])$                &  $(k+2, H, W)$  \\  \hline                             
			\textit{ResNet-34 backbone + ReLU}             & $(k+2, H, W)$         & $(512, H//32, W//32)$   \\  \hline
			\textit{Fully connected + ReLU}             & $(512 \times H//32 \times W//32)$                &  $(32 \times \left\lfloor \frac{H}{a} \right\rfloor \times \left\lfloor \frac{W}{2b} \right\rfloor)$   \\ \hline 
			\textit{Fully connected + ReLU}             & $(32 \times \left\lfloor \frac{H}{a} \right\rfloor \times \left\lfloor \frac{W}{2b} \right\rfloor)$                 &  $(16 \times \left\lfloor \frac{H}{a} \right\rfloor \times \left\lfloor \frac{W}{2b} \right\rfloor)$   \\ \hline 
			\textit{Conv + Sigmoid + flatten}             & $(16, \left\lfloor \frac{H}{a} \right\rfloor, \left\lfloor \frac{W}{2b} \right\rfloor)$                 &  $(\left\lfloor \frac{H}{a} \right\rfloor \times \left\lfloor \frac{W}{2b} \right\rfloor)$   \\
			\bottomrule
		\end{tabular}
		\vspace{+3mm}
		\caption{\label{tab:approx_nn_arch} NN-based reachability approximator model input, architecture, and output.}
\end{table}
The model of the NN-based reachability approximator uses the Resnet-34 as a backbone and a customized output head to output the reachability indicator vector. \cref{tab:approx_nn_arch} lists the input, output, and the model architecture. The inputs of the model are $\mathbf{I}_{t:t+k}$ and $\mathbf{e}_t$ (we pre-process $\mathbf{e}_t$ using a fully connected layer to transform it into same shape of the input BEV images as an additional channel), the output of the model is a vector of size $\left\lfloor \frac{H}{a} \right\rfloor \times \left\lfloor \frac{W}{2b} \right\rfloor$. We omit the mini-batch dimension, and the usage of $flatten$ when appropriate for better clarity.

\section{SMC Additional Details}
\subsection{Mutated scenarios hyperparameters \label{sm:scene_param}}
The 1000 mutated scenarios are generated based on the cut in safety-critical scenario typology, as described in \cref{s:datasets}, by varying multiple adjustable hyperparameters as follows.
\begin{enumerate}
	\item \textit{Distance traveled before cut in}: This parameter specifies the distance that the NPC actor travels before starting to cut in. The lower the value, the sooner the cut in happens after the NPC actor passes the ego actor, the more severe the criticality. We vary this parameter value in the range of $[10, 20)$, with an increment of 1. 
	\item \textit{Distance traveled during lane change}: This parameter specifies the distance that the NPC actor travels during the cut in. The lower the value, the steeper the cut in angle, the more severe the criticality, and vise versa. We vary this parameter value in the range of $[6, 16)$, with an increment of 1. 
	\item \textit{Lane change speed}: This parameter specifies the speed of the NPC actor during the cut in. The lower the value, the lower the cut in speed, the longer it takes to complete the cut in, and vise versa. 
	We vary this parameter value in the range of $[9, 19)$, with an increment of 1. 
\end{enumerate}
We then generate scenarios using all combinations of these three hyperparameters, which in term resulted in 1000 mutations of the base scenario. 
For each run of the \textit{LBC} agent on these mutated scenarios, we collect the simulator data on actors and the map information which results in a dataset similar to the \textit{Argoverse} dataset.
This dataset is referred to as \textit{mutated scenarios dataset}. 

We can convert this dataset using the same method outlined in \cref{s:nn_reach} and \cref{sm:data_gen} in to the training and evaluation dataset for the NN-based reachability approximator to be used in simulator environment. The training dataset is generated from 80\% of the 1000 scenarios which consists of $\sim 340$K datapoints. The testing dataset is generated from the rest 20\% which consists of $\sim 83$K datapoints. 
Additionally, we perform a 80\%-20\% split on the training data set for validation during the training procedure.

\paragraph{Training and evaluating the SMC.}
We also use one (the scenario with the median value of individual hyperparameter ranges) of these 1000 mutated scenarios to train the SMC and learn a mitigation policy (\cref{s:smc_training}). We then evaluate this policy using all of the 1000 mutated scenarios to assess the generalizability of the learned policy. The result in \cref{s:smc_results} shows that the policy is generalizable to different level of safety criticality of the similar scenario typology as SMC reduces the accident rate significantly.

\subsection{RL Reward function}
\cref{ml4ad:eq:rl_eward} defines a general reward formulation as a multi-parts weight sum to encode the trade-offs of mitigation actions. This subsection defines the various terms used in this reward function. In general, the reward function should encourage mitigation in high \score situation, but discourage mitigation that affects normal driving in low \score situation.
\begin{enumerate}
	\item \textit{\score}: The value of the scene \score evaluated per simulation time step. The term $(1-\score)$ encourages low \score and discourage high \score. During our implementation, we do find that this term dominates and continue to provide large positive rewards even if the mitigation action like emergency braking is activated continuously, as long as the scene risk is 0 or low. This lead to a mode collapse as the ego vehicle continues to receive reward even if it is stationary (holding the emergency brake). We reduce this effect by multiplying this term with the speed factor $v_f$ (described as part of the $r_{pc}$): $(1-\score)v_f$ so that no rewards are given if the ego actor is stationary.
	\item \textit{$r_{pc}$}: The path-completion reward term describes the reward for completing the path.
	\begin{equation}
		r_{pa} = (\mathbf{u}_{ego} \cdot \mathbf{u}_{goal}) v_f
	\end{equation}
	\begin{equation}
		v_f = \frac{\|\mathbf{d}_{ego}\|}{d_{expected}} 
	\end{equation}
	where $\mathbf{u}_{ego}$ is the unit vector of the ego actor's traveling direction, $\mathbf{u}_{goal}$ is the unit vector of the goal/next-planned waypoint direction, $\|\mathbf{d}_{ego}\|$ is the distance covered by ego actor per time step, and $d_{expected}$ is the expected coverage if the road is NPC actor free. $r_{pc}$ encourages 
	the actor to travel in the planned direction with the expected speed, and discourages large deviation from the planned route or the expected speed.
	\item \textit{$p_{am}$}: The active-mitigation penalty term penalizes the activation of the mitigation action in general because mitigation actions can change the dynamic of the ego actor drastically, resulting in passenger discomfort and even panics. This term can be a constant that applies whenever mitigation is activated, and otherwise zero, we do found that introduces such a term lead to discontinuity in the reward function which affects the Q-function's convergence, so a smoother function is needed. As a result we set this term to zero in our experiments.
	\item \textit{$r_{comfort}$}: Given that the ego agents used/evaluated in this paper~\cite{chen2019lbc,pmlr-v119-filos20a} do not account for comfort explicitly, we also decided to not account for comfort to reduce the time for training. However, the reward model can be extended to account for comfort by using equation \cref{sm_eq:comfort}. 
	In real world modeling comfort is important because an aggressive mitigation policy can cause unnecessary discomforts. Comfort is commonly characterized as the rate of change in acceleration so one way to define $r_{comfort}$ is as follows, where $a$ is the acceleration:
	\begin{equation}
		r_{comfort} =  -(j_{lat} + j_{long}), j = \frac{da}{dt}
		\label{sm_eq:comfort}
	\end{equation}
\end{enumerate}

\subsection{SMC model architecture}
\label{sm:smc_ma}
As specified in \cref{s:exp_smc} the SMC's Q-value function uses a similar architecture as the \textit{LBC} agent's \textit{Sensorimotor model} \cite{chen2019lbc} with the only changes being the input and the output format. We choose this design as the SMC consumes similar input as the original ego actor's input to more efficiently reuse information and reduce complexity. 

We change the original input format, a $10$-channel 3D feature maps that consists of 3 RGB images from 3 front facing cameras and 1 target heatmap, to a $16$-channel 3D feature maps. Instead of using 3 RGB images as part of the input, we convert these 3 RGB images into 3 grayscale images. In addition, we collect such feature maps for 4 consecutive simulation time step to capture the temporal information, so the resulting feature maps is consists of 16 channels, with the first 12 channels being the grayscale images (3 cameras \texttimes 4 time steps), and 4 target heatmap (1 heatmap per time step). 

The output format of the original \textit{Sensorimotor model} is a $4 \times 2$ vector that contains $4$ local waypoints to follow. We change this format and now the output equals to the dimension of the action space with each element specify the Q-value of that action; in our case, the output is $1 \times 2$.

We refer the reader to \cite{chen2019lbc} for more details of the architecture used.
\section{Compute Hardware and Compute Hours}
\subsection{Hardware specifications}
We train and evaluate the NN-based reachability approximator and the RL-based SMC on a system with AMD Ryzen 9 3950X CPU, 64 GB of system memory, and a RTX 3090 TI with 24 GB of video memory. For training and evaluation dataset label generation and dataset \score characterization using the \textit{FOT} planner based \score evaluator, a more powerful machine with AMD Ryzen Threadripper 3990X CPU and 128 GB of system memory is used to reduce computation time.

\subsection{Compute hours}
The NN-based approximator training and evaluation dataset generation from the $Arogoverse$ real-world dataset takes 25 CPU hours using the \textit{FOT} planner based \score evaluator to generate label. The generation of dataset from the 1000 mutated scenarios takes 25 + 27 CPU hours, with the first 25 hours used for running and recording all 1000 mutated scenarios using the LBC agent, and the next 27 hours for label generation using the \textit{FOT} planner based \score evaluator. Training of the NN-based reachability approximator for 500 epochs requires 9 hours of GPU time on the \textit{mutated scenarios dataset} and 7 hours of GPU time on the \textit{Argoverse dataset}. The evaluation of NN-based reachability approximator on real world dataset consumes 1 hours of GPU time, while the evaluation on the simulated dataset consumes 2 hours due to more data.
Training of the RL-based SMC for 100 episodes requires roughly 500 minutes or 8.3 hours, with each episode consumes approximately 5 minutes of wall clock time. The evaluation of the SMC on the 1000 mutated scenarios takes around 33 hours with each run finishes in 2 minutes. For baseline comparison, evaluation of the \textit{LBC} agent on the 1000 mutated scenarios takes around the same time as with SMC enabled. For comparison with the \textit{RIP} agent, evaluation on the 1000 mutated scenarios takes around 50 hours, with each run finishes in 3 minutes.

\section{Hyperparameters}
We list the hyperparameters used in this work below. 
\begin{enumerate}
	\item \textit{FOT} planner based \score calculation with BEV representation (\cref{s:reach_cal_fp,sm:bev_rep}).
	\begin{itemize}
		\item Time budget (Look forward time): $3$ seconds is equivalent to $k=30$ as all data (real and simulated) are collected at 10 Hz.
		\item Distance budget (Look forward distance): $d_{max} = 120$ meters, distance threshold for potential local goals is $d_{thres} = \max(d, d_{max})$, where $d$ is specified in \cref{sm_eq:d}.
		\item Lane width (grid width): $w = 3.7$ meters for \textit{Argoverse dataset}; width varies with \textit{mutated scenarios dataset} dataset based on actual simulated lane width. 
		\item Grid distance (grid length): $l=4.5$ meters for both \textit{Argoverse dataset} and \textit{mutated scenarios dataset}.
		\item Minimum distance to static obstacles (lane boundaries): 0.1 meters.
		\item Minimum distance to dynamic obstacles (NPC actors): 1.5 meters.
		\item \textit{FOT} planner hyperparameters based on typical consumer vehicle.
		\begin{enumerate}
			\item Maximum speed: 100 $km/h$, 27.7 $m/s$.
			\item Maximum acceleration: 4.0 $m/s^2$.
			\item Maximum turning curvature: 0.2 $m^{-1}$.
		\end{enumerate}
	\end{itemize}
	\item NN-based reachability approximation.
	\begin{itemize}
		\item Reachability approximation dataset generation.
		\begin{enumerate}
			\item All hyperparameters used in \textit{FOT} planner based \score calculation with BEV representation for reachability label generation.
			\item BEV image height: $H=128$ pixels.
			\item BEV image width: $W=512$ pixels.
			\item X-axis view range (x-axis): $R_X=17.5$ meters. 
			\item Y-axis view range (y-axis): $R_Y=70.0$ meters.
			\item \# of pixels in the reachability grid in the X-direction: $a=11$ pixels.
			\item \# of pixels in the reachability grid in the Y-direction: $b=20$ pixels.
		\end{enumerate}
		\item Model hyperparameters: The hyperparameters for the NN-based reachability approximator's neural network model can be derived from the shape of the input and output specified in \cref{tab:approx_nn_arch}.
		\item Training and validation.
		\begin{enumerate}
			\item Adam optimizer learning rate: \num{1e-5}.
			\item Adam optimizer weight decay: $0.99$.
			\item Number of epoch: $500$.
		\end{enumerate}
	\end{itemize}
	\item RL-based SMC for mitigation.
	\begin{itemize}
		\item Mutated scenarios dataset generation: See \cref{sm:scene_param} for hyperparameters used.
		\item Model hyperparameters: The SMC Q-value model's neural network model are specified in \cref{sm:smc_ma} and \cite{chen2019lbc}.
		\item Training and evaluation.
		\begin{enumerate}
			\item Adam optimizer learning rate: \num{1e-5}.
			\item Adam optimizer weight decay: $0.9999$.
			\item RL reward decay: $0.99$.
			\item Number of episodes: $100$.
		\end{enumerate}
	\end{itemize}
\end{enumerate}

\end{document}